\documentclass{article} % For LaTeX2e
\usepackage{iclr2023_conference,times}

\usepackage{amsmath,amsfonts,bm}

\def\Figref#1{Figure~\ref{#1}}

\def\Secref#1{\S~\ref{#1}}
\def\eqref#1{equation~\ref{#1}}
\def\1{\bm{1}}

\DeclareMathAlphabet{\mathsfit}{\encodingdefault}{\sfdefault}{m}{sl}
\SetMathAlphabet{\mathsfit}{bold}{\encodingdefault}{\sfdefault}{bx}{n}

\usepackage{multirow}
\usepackage{makecell}
\usepackage{threeparttable}
\usepackage{booktabs}

\usepackage{url}
\usepackage{tablefootnote}
\usepackage{listings}
\usepackage[pdftex]{graphicx}

\definecolor{airforceblue}{rgb}{0.36, 0.54, 0.66}
\definecolor{bluegray}{rgb}{0.4, 0.6, 0.8}
\definecolor{bleudefrance}{rgb}{0.19, 0.55, 0.91}
\usepackage[colorlinks=true,linkcolor=orange,citecolor=bleudefrance]{hyperref}
\newcommand{\tabref}[1]{Table~\ref{#1}}

\title{DiffuSeq: Sequence to Sequence \\Text Generation with Diffusion Models}

\author{Shansan Gong$^1$, Mukai Li$^1$, Jiangtao Feng$^1$, Zhiyong Wu$^1$, Lingpeng Kong$^{2}$\\
$^1$Shark-NLP, Shanghai AI Laboratory \quad $^2$The University of Hong Kong\\
\texttt{\{gongshansan,limukai,fengjiangtao,wuzhiyong\}@pjlab.org.cn}\\
\texttt{lpk@cs.hku.hk}
} 

\usepackage{xspace}
\iclrfinalcopy % Uncomment for camera-ready version, but NOT for submission.
\begin{document}

\newcommand{\model}{\textsc{DiffuSeq}\xspace}

\maketitle

\begin{abstract}
Recently, diffusion models have emerged as a new paradigm for generative models. Despite the success in domains using continuous signals such as vision and audio, adapting diffusion models to natural language is under-explored due to the discrete nature of texts, especially for conditional generation. We tackle this challenge by proposing \model: a diffusion model designed for sequence-to-sequence (\textsc{Seq2Seq}) text generation tasks. 
Upon extensive evaluation over a wide range of \textsc{Seq2Seq} tasks, we find \model achieving comparable or even better performance than six established baselines, including a state-of-the-art model that is based on pre-trained language models. 
Apart from quality, an intriguing property of \model is its high diversity during generation, which is desired in many \textsc{Seq2Seq} tasks.  We further include a theoretical analysis revealing the connection between \model and autoregressive/non-autoregressive models. Bringing together theoretical analysis and empirical evidence, we demonstrate the great potential of diffusion models in complex conditional language generation tasks. 
\footnote{Code is available at \url{https://github.com/Shark-NLP/DiffuSeq}}

\end{abstract}

\section{Introduction}
Among existing generative models, GAN~\citep{NIPS2014_5ca3e9b1} suffers from the instability issue~\citep{NIPS2016_8a3363ab}, subjecting to mode collapse~\citep{iclrMetzPPS17}; VAE~\citep{Kingma2014} has to rely on surrogate objectives to approximate maximum likelihood training and Flow-based models~\citep{iclrDinhSB17} has to use specialized architectures to construct reversible transform. Diffusion models~\citep{ho2020denoising, nichol2021improved} have circumvented several of these limitations and emerged as a new paradigm for generative models, theoretically underpinned by non-equilibrium thermodynamics~\citep{sohl2015deep} and score-matching network~\citep{song2019generative}. To date, the major breakthroughs are in domains using continuous signals, such as vision~\citep{saharia2022palette,saharia2022photorealistic,ramesh2022hierarchical} and audio~\citep{kong2020diffwave}. However, extending continuous diffusion models to natural language remains an open challenge due to the inherently discrete nature of texts.

\begin{figure}[ht]
  \centering
  \includegraphics[width=\linewidth]{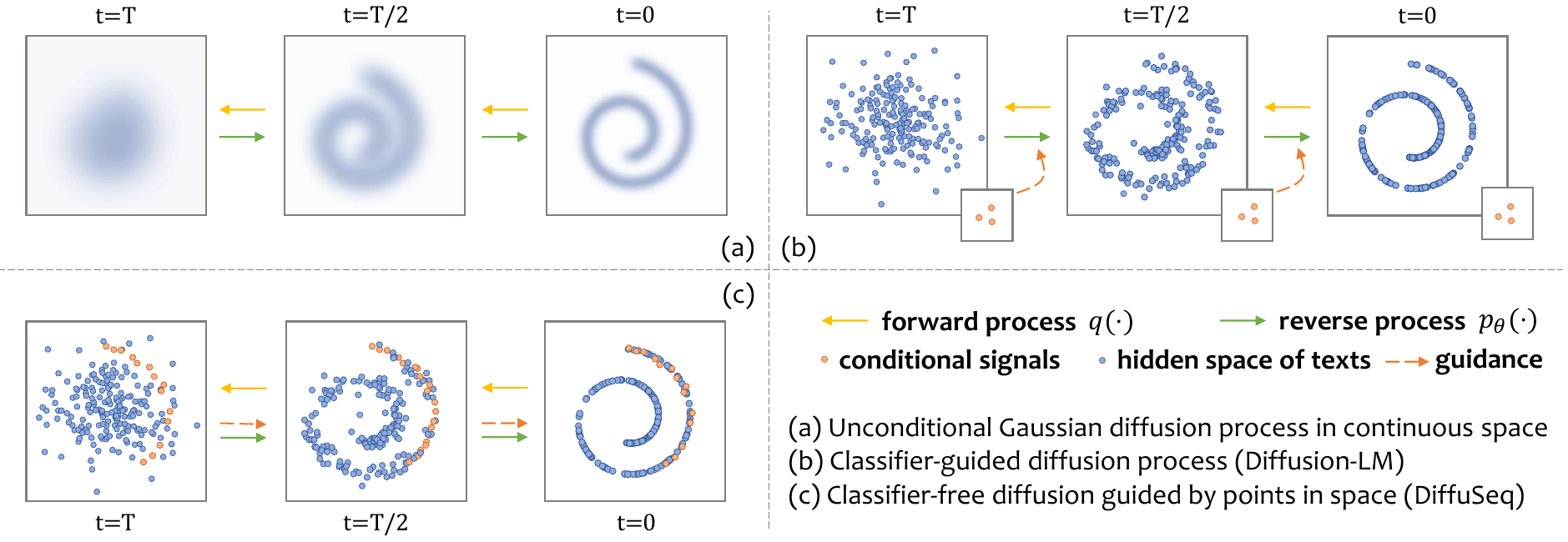}
  \caption{The demonstration of unconditional, classifier-guided, and classifier-free diffusion models.}
  \label{fig:intro}
\end{figure}

On the basis of unconditional generation in continuous space which is illustrated in~\Figref{fig:intro}(a), existing efforts~\citep{hoogeboom2021argmax,austin2021structured} start customizing diffusion models to text in discrete space on unconditional language modeling (i.e., free text generation).
Diffusion-LM~\citep{li2022diffusion}, as in~\Figref{fig:intro}(b), models texts in continuous space and proposes to use an extra-trained classifier as guidance (i.e., the condition signal $\textbf{x}$) to impose subtle changes (usually complex, fine-grained constraints) on generated sentences. Nonetheless, these models do not naturally generalize to conditional language modeling (i.e., the model assigns probabilities $p(\textbf{w}|\textbf{x})$ to sequences of words $\textbf{w}$ given $\textbf{x}$). In the more general sequence-to-sequence (\textsc{Seq2Seq}) setting where the condition $\textbf{x}$ is also a sequence of words, applying Diffusion-LM can be difficult. The reason is that classifiers are attributes-oriented, and we can not train hundreds-of-thousands classifiers to model the semantic meaning between conditions and generated sentences.

\textsc{Seq2Seq} is an essential setting in NLP that covers a wide range of important tasks such as open-ended sentence generation, dialogue, paraphrasing, and text style transfer. In this paper, we propose \model, depicted in~\Figref{fig:intro}(c), a classifier-free diffusion model that supports \textsc{Seq2Seq} text generation tasks. By modeling the conditional probability of the target sentence $\textbf{w}$ given context $\textbf{x}$ using one single model, one advantage of \model is that this paradigm allows a complete model to fit data distribution and utilize conditional guidance, rather than depending on a separate classifier. 

Different from canonical generation approaches in an autoregressive (AR) left-to-right manner~\citep{radford2019language}, \model generates text tokens parallelly in the non-autoregressive (NAR) way. To corroborate the effectiveness of our \textsc{DiffuSeq}, we conduct experiments on four \textsc{Seq2Seq} tasks. Compared to AR and NAR models, which suffer from the ``degeneration'' problem~\citep{holtzman2019curious} and rely on decoding strategies, \textsc{DiffuSeq} can achieve considerable sentence-level diversity without sacrificing the quality (see~\Secref{sec:result}).

To sum up, we make a series of technical and conceptual contributions:
(a) we are the first to deploy the diffusion model on \textsc{Seq2Seq} text generation, and our proposed \model as a conditional language model is trained end-to-end in a classifier-free manner; 
(b) we establish a theoretical connection among AR, NAR and \model models, and justify \model as an extension of iterative-NAR models;
(c) with strong empirical evidence, we demonstrate the great potential of diffusion models in complex conditional language generation tasks.
\section{Preliminary and Problem Statement}
% \subsection{Preliminary and Problem Statement}
\label{sec:prel}
\paragraph{Preliminary.} 
A diffusion model typically contains forward and reverse processes. 
Given a data point sampled from a real-world data distribution $\mathbf{z}_0 \sim q(\mathbf{z})$, the forward process gradually corrupts $\mathbf{z}_0$ into a standard Gaussian noise $\mathbf{z}_T \sim \mathcal{N}(0, \mathbf{I})$. For each forward step $t \in [1, 2,...,T]$, the perturbation is controlled by $q(\mathbf{z}_{t} \vert \mathbf{z}_{t-1}) = \mathcal{N}(\mathbf{z}_{t};\sqrt{1-\beta_t}\mathbf{z}_{t-1}, {\beta}_t \mathbf{I})$, with $\beta_t \in (0,1)$ as different variance scales. Once the forward process is completed, the reverse denoising process tries to gradually reconstruct the original data $\mathbf{z}_0$ via sampling from $\mathbf{z}_T$ by learning a diffusion model $f_{\theta}$. 
% The model is optimized via the variational lower bound on negative log-likelihood: $\mathbb{E}[-\log p_{\theta}(\mathbf{x}_0) ]\leq L_\text{VLB}$.
% \paragraph{Diffuse Text in Continuous Space}
% To adapt conventional diffusion models to text, Diffusion-LM~\citep{li2022diffusion} introduced following techniques. First, they use an embedding function $\textsc{Emb}(\mathbf{w})$ to map the discrete text $\mathbf{w}$ into a continuous space $\textbf{x}_0$, as $q_{\phi}(\mathbf{x}_0|\mathbf{w})=\mathcal{N}(\textsc{Emb}(\mathbf{w}), \beta_0 \mathbf{I})$. This transformation allows us to incorporate text into the standard forward process. \lpk{left here. come back later.}
% Second, during decoding, they design a clamping method to map the perturbed embedding vectors back to words.  
\paragraph{Problem Statement.} Many recent efforts have been devoted to adapting diffusion models to discrete texts (See \Secref{sec:related}). However, they all focus on unconditional sequence modeling. 
In this paper, we target the sequence-to-sequence text generation tasks. In particular, given a $m$-length source sequence $\mathbf{w}^x = \{w^x_1, ..., w^x_m\}$, we aim to learn a diffusion model that can produce a $n$-length target sequence $\mathbf{w}^y = \{w^y_1, ..., w^y_n\}$ conditioning on the source sequence.

\section{\model}

% In sequence to sequence tasks, given two sequences with the source document of length $n$ as $\mathbf{w}^x = \{w^x_1, ..., w^x_n\}$ and the target document of length $m$ as $\mathbf{w}^y = \{w^y_1, ..., w^y_m\}$, we aim to train a diffusion model to achieve the conversion from the discrete word space $\mathbf{w}^x$ to $\mathbf{w}^y$. We denote the pair-wisely concatenated space of sequence $\mathbf{w}^x$ and $\mathbf{w}^y$ during the diffusion process as $\mathbf{z}_t = \mathbf{x}_t \oplus \mathbf{y}_t$. 

\begin{figure}[t]
  \centering
  \includegraphics[width=0.99\linewidth]{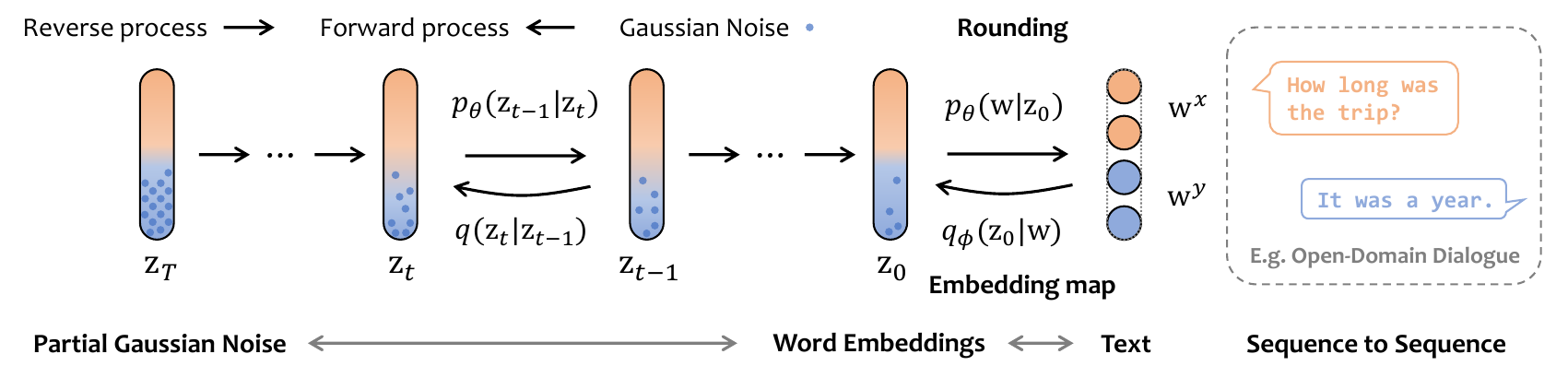}
  \caption{The diffusion process of our conditional diffusion language model \textsc{DiffuSeq}. Given the source $\mathbf{w}^x$ and the target $\mathbf{w}^y$, we pair-wisely transform them into continuous space $\mathbf{z}_0$. The partial Gaussian noise is iteratively added on the target space of $\mathbf{z}_t$.}
  \label{fig:seq2seq}
\end{figure}

% Referring to conditional-VAE~\citep{zhao2017learning}, we can consider the latent encoded input $\mathbf{x}$ as a condition. 

% Constructing models to support \textsc{Seq2Seq} tasks requires several modifications to the vanilla diffusion model. 
We propose \model to extend vanilla diffusion models to learn conditional text generation~(as shown in~\Figref{fig:seq2seq}), concerning the model architecture and the training objective. 

\paragraph{Forward Process with Partial Noising.} In the beginning of forward process, we follow Diffusion-LM~\citep{li2022diffusion} to design an embedding function $\textsc{Emb}(\mathbf{w})$ to map the discrete text $\mathbf{w}$ into a continuous space.
% $\textbf{z}_0$
% ~\footnote{We use $z$ to denote the concatenated diffusion hidden state and distinguish from $x$ used in \S~\ref{sec:prel}}. 
% In particular, given a pair of sequences $\mathbf{w}^x$ and $\mathbf{w}^y$, we first concatenate them to get a combined sequence $\mathbf{w}$ and then obtain the continuous representation as $\textsc{Emb}(\mathbf{w}^{x\oplus y}) = [\textsc{Emb}(w^x_1),...,\textsc{Emb}(w^x_n), \textsc{Emb}(w^y_1),...,\textsc{Emb}(w^y_m)] \in \mathbb{R}^{(n+m)\times d}$.
In particular, given a pair of sequence $\mathbf{w}^x$ and $\mathbf{w}^y$, \model learns a unified feature space of $\mathbf{w}^x$ and $\mathbf{w}^y$ by embedding transformation and concatenation as $\textsc{Emb}(\mathbf{w}^{x\oplus y}) = [\textsc{Emb}(w^x_1),...,\textsc{Emb}(w^x_m), \textsc{Emb}(w^y_1),...,\textsc{Emb}(w^y_n)] \in \mathbb{R}^{(m+n)\times d}$.
The transformation allows us to adapt discrete textual input into the standard forward process, by extending the original forward chain to a new Markov transition ${q_{\phi}}(\mathbf{z}_{0}|\mathbf{w}^{x\oplus y})=\mathcal{N}(\textsc{Emb}(\mathbf{w}^{x\oplus y}), \beta_0 \mathbf{I})$. 

% The ultimate goal of the forward process is to transform $z_0$ into a standard Gaussian noise $\mathbf{z}_T \sim \mathcal{N}(0, \mathbf{I})$, denoting as $ q(\mathbf{z}_{T}):=\prod_{t=1}^Tq(\mathbf{z}_t|\mathbf{z}_{t-1})$. 
We denote $\mathbf{z}_{t} = \mathbf{x}_t \oplus \mathbf{y}_t$ to simplify the wordings, where $\mathbf{x}_t$ and $\mathbf{y}_t$ represent parts of $\mathbf{z}_{t}$ that belong to $\mathbf{w}^x$ and $\mathbf{w}^y$, respectively. For each forward step $ q(\mathbf{z}_t|\mathbf{z}_{t-1})$, we gradually inject noise into last step's hidden state $\mathbf{z}_{t-1}$ to obtain $\mathbf{z}_{t}$. Unlike conventional diffusion models that corrupt the whole $\mathbf{z}_{t}$ (both $\mathbf{x}_t$ and $\mathbf{y}_t$) without distinction, we only impose noising on $\mathbf{y}_t$. This modification (termed \textbf{{partial} noising}) allows us to {adapt} diffusion models {for} conditional language modeling.

\paragraph{Reverse Process with Conditional Denoising.}
The ultimate goal of the reverse process is to recover the original $\mathbf{z}_0$ by denoising $\mathbf{z}_t$: $p_{\theta}(\mathbf{z}_{0:T}):=p(\mathbf{z}_T)\prod_{t=1}^Tp_{\theta}(\mathbf{z}_{t-1}|\mathbf{z}_t)$. We model the learning process $p_{\theta}(\mathbf{z}_{t-1}|\mathbf{z}_t)=\mathcal{N}(\mathbf{z}_{t-1};\mu_{\theta}(\mathbf{z}_t, t), \sigma_{\theta}(\mathbf{z}_t, t))$ using the proposed diffusion model $\textsc{DiffuSeq}$: $f_{\theta}(\mathbf{z}_t, t)$, where the $\mu_{\theta}(\cdot)$ and $\sigma_{\theta}(\cdot)$ is the parameterization of the predicted mean and standard {deviation} of $ q(\mathbf{z}_{t-1}|\mathbf{z}_{t})$ in forward process, derived using Bayes' rule. The detailed derivations are in Appendix~\ref{sec:appendix-obj}. With the partial nosing strategy adopted in the forward process, we can impose the input as the condition when denoising as shown in~\Figref{fig:intro}. 
The proposed conditional denoising is classifier-free by nature: we do not require extra-trained classifiers to control the denoising process. 

Specifically, we use a transformer architecture to model $f_{\theta}$, which spontaneously models the semantic relation between $\mathbf{x}_t$ and $\mathbf{y}_t$. We compute the variational lower bound ($\mathcal{L}_\text{VLB}$) following the original diffusion process. {$\mathcal{L}_{round}$ corresponds to rounding operation in~\Figref{fig:seq2seq}.}

% \paragraph{Conditional Reverse Noising}
% The reverse process then denoises $\mathbf{z}_t$, aiming to recover original $\mathbf{z}_0$, and is defined as: $p_{\theta}(\mathbf{z}_{0:T}):=p(\mathbf{z}_T)\prod_{t=1}^Tp_{\theta}(\mathbf{z}_{t-1}|\mathbf{z}_t)$. The learning of $p_{\theta}(\mathbf{z}_{t-1}|\mathbf{z}_t)=\mathcal{N}(\mathbf{z}_{t-1};\mu_{\theta}(\mathbf{z}_t, t), \sigma_{\theta}(\mathbf{z}_t, t))$ is based on our diffusion model $\textsc{DiffuSeq}$: $f_{\theta}(\mathbf{z}_t, t)$, where the $\mu_{\theta}(\cdot)$ and $\sigma_{\theta}(\cdot)$ is the parameterization of the mean and standard variation of $\tilde q(\mathbf{z}_t|\mathbf{z}_{t-1})$ in forward process.
% Specifically, we can use a transformer architecture to model $f_{\theta}$, which spontaneously models the attention relation between $\mathbf{x}_t$ and $\mathbf{y}_t$ and is easily plugged with the time step embedding referring to the position embedding. We can compute the Variational Lower Bound following the original diffusion theory. (Details in Appendix~\ref{sec:appendix-obj}.)

\begin{equation}
\begin{aligned}
\mathcal{L}_\text{VLB}
% L_T + L_{T-1} + \dots + L_0
&=\mathbb{E}_{ q(\mathbf{z}_{1:T}|\mathbf{z}_0)}
\Bigg[
\underbrace{\log\frac{ q(\mathbf{z}_T|\mathbf{z}_0)}{p_{\theta}(\mathbf{z}_T)}}_{\mathcal{L}_T}  + \sum_{t=2}^T \underbrace{\log{\frac{ q(\mathbf{z}_{t-1}|\mathbf{z}_0,\mathbf{z}_t)}{p_{\theta}(\mathbf{z}_{t-1}|\mathbf{z}_t)}}}_{\mathcal{L}_{t-1}} \\
& + {\underbrace{\log\frac{ q_{\phi}(\mathbf{z}_0|\mathbf{w}^{x\oplus y})}{ p_{\theta}(\mathbf{z}_0|\mathbf{z}_1)}}_{\mathcal{L}_0}}-\underbrace{\log p_{\theta}(\mathbf{w}^{x\oplus y}|\mathbf{z}_0)\vphantom{\log{\frac{ q(\mathbf{z}_{t-1}|\mathbf{z}_0,\mathbf{z}_t)}{p_{\theta}(\mathbf{z}_{t-1}|\mathbf{z}_t)}}}}_{\mathcal{L}_\text{round}}
\Bigg].\\
\end{aligned}
\end{equation}
% For $1 \leq t \leq T-1$, we compute the parameterization of $L_t$ to minimize the difference from $\mu$ and $\mu_{\theta}$ following~\citet{ho2020denoising}.
% \begin{equation}
% \begin{aligned}
% L_t&=\mathbb{E}_{\mathbf{z}_0}\left[\log{\frac{\tilde q(\mathbf{z}_{t}|\mathbf{z}_0,\mathbf{z}_{t+1})}{p_{\theta}(\mathbf{z}_{t}|\mathbf{z}_{t+1})}}\right]= \mathbb{E}_{\mathbf{z}_0}\left[\frac{1}{\mathcal{C}}|| \mu_t(\mathbf{z}_t,\mathbf{z}_0)-\mu_{\theta}(\mathbf{z}_t, t)||^2 \right]\\ & =\mathbb{E}_{\mathbf{z}_0}\left[\frac{1}{\mathcal{C}}||\mathcal{A}\mathbf{z}_t+\mathcal{B}\mathbf{z}_0-(\mathcal{A}\mathbf{z}_t + \mathcal{B}f_{\theta}(\mathbf{z}_t, t))||^2\right]=\frac{\mathcal{B}}{\mathcal{C}}\mathbb{E}_{\mathbf{z}_0}[||\mathbf{z}_0-f_{\theta}(\mathbf{z}_t, t)||^2],  \\
% \end{aligned}
% \end{equation}
% where $\mathcal{C}=2||\sigma_{\theta}||^2$, $\mathcal{A}=\frac{\sqrt{\alpha_t}(1-\bar{\alpha}_{t-1})}{1-\bar{\alpha}_t}$, and $\mathcal{B}=\frac{\sqrt{\bar{\alpha}_{t-1}}\beta_t}{1-\bar{\alpha}_t}$, with $\alpha_t=1-\beta_t$ and $\bar{\alpha}_t=\prod_{i=1}^t\alpha_i$. 
We further simplify the training objective as follows (details in Appendix~\ref{sec:appendix-obj}):
$$\min_{\theta}\; \mathcal{L}_{\text{VLB}} = \min_{\theta}\left[
% ||\mu(\mathbf{z}_T)||^2+
\sum_{t=2}^T||\mathbf{z}_0-f_{\theta}(\mathbf{z}_t, t)||^2 + ||\textsc{Emb}(\mathbf{w}^{x\oplus y})-f_{\theta}(\mathbf{z}_1, 1)||^2-\log p_{\theta}(\mathbf{w}^{x\oplus y}|\mathbf{z}_0)\right] $$
\begin{equation}
\label{eq:loss}
\rightarrow \min_{\theta}\left[ \sum_{t=2}^T||\mathbf{y}_0-\tilde f_{\theta}(\mathbf{z}_t, t)||^2 + {||\textsc{Emb}(\mathbf{w}^y)-\tilde f_{\theta}(\mathbf{z}_1, 1)||^2} + \mathcal{R}(||\mathbf{z}_0||^2)\right],
% \end{aligned}
\end{equation}
here we use $\tilde f_{\theta}(\mathbf{z}_t, t)$ to denote the fractions of recovered $\mathbf{z}_0$ corresponding to $\mathbf{y}_0$. Note that although in the first term, we only compute the loss w.r.t $\mathbf{y}_0$, due to the attention mechanism in the transformer, the reconstruction of $\mathbf{y}_0$ also takes $\mathbf{x}_0$ into account, thus the gradients from the first term will also affect the learning of $\mathbf{x}_0$. {The mathematically equivalent regularization term $\mathcal{R}(||\mathbf{z}_0||^2)$) regularize the embedding learning.} We further share the embedding function between source and target sequences, enabling the training of two different feature spaces jointly. This sets \textsc{DiffuSeq} away from existing solutions in vision such as GLIDE~\citep{nichol2021glide}.

% but the reconstruct of $\mathbf{y}_0$ is associated with $\mathbf{x}_0$ in the nn model, with additional regularization $\mathcal{R}$ of $\textbf{x}_0$.

\paragraph{Training and Inference Methods.}
% During the training process, the defined embeddings of corresponding discrete words is 
% ly trained in~\Eqref{eq:loss}. Due to the clamping-like operation in $\mathbf{x}$ space of $\mathbf{z}$, the $\mathbf{x}_t$ is consistently replaced by $\mathbf{x}_0$, but the condition $\mathbf{x}$ still takes its effect in two ways. \TODO{explain the embedding of x}
In our preliminary experiments, we find that the high diversity in NLP datasets and long diffusion steps often result in insufficient training. We hypothesize the reason is that sampling step $t$ uniformly causes unnecessary noise in the $\mathcal{L}_{\text{VLB}}$ objective. 
% To reduce the out-of-vocabulary (OOV) generation and improve the efficiency of vocabulary learning, we apply Byte Pair Encoding~\citep{sennrich-etal-2016-neural} to construct the vocabulary. A larger vocabulary, however,  and it empirically brings more burden for model training. 
We hence employ importance sampling~\citep{nichol2021improved} to address this problem.
\begin{equation}
    \mathcal{L}_{\text{VLB}} = \mathbb{E}_{t\sim p_t}\left[\frac{\mathcal{L}_t}{p_t}\right], \;p_t \propto \sqrt{\mathbb{E}[\mathcal{L}_t^2]},\;\textstyle\sum_{t=0}^{T-1} p_t=1.
\end{equation}

Intuitively, the importance-weighted sampling algorithm will spend more steps on diffusion steps with larger $\mathcal{L}_t$, and vice versa. 
% This dynamic training algorithm 
% \paragraph{Decoding}
% \paragraph{Conditional inference}

To conduct \textsc{Seq2Seq} generation given the condition $\textsc{Emb}(\mathbf{w}^x)$, we randomly sample $\mathbf{y}_T\sim\mathcal{N}(0, I)$ and  concatenate $\mathbf{y}_T$ with $\textsc{Emb}(\mathbf{w}^x)$ to obtain $\mathbf{z}_T$. We can now repeat the reverse process until we arrive at $\mathbf{z}_0$. At each sampling step, an anchoring function is executed towards reparameterized $\mathbf{z}_t$. Specifically, the anchoring function: (a) operates rounding on $\mathbf{z}_t$ to map it back to word embedding space following~\citet{li2022diffusion}; (b) replaces the part of recovered $\mathbf{z}_{t-1}$ that belongs to $\mathbf{w}^x$ with the original $\mathbf{x}_0$, considering that this part is recovered from corrupted $\mathbf{z}_t$ via $f_\theta$ and not strictly equals to $\mathbf{x}_0$. Note that (b) is designed for \model.

% When sampling the generated responses using $p_{\theta}(\mathbf{z}_{t-1}|\mathbf{z}_t)$, we typically follow the reverse sequence of time steps $T,...,2,1$. The initial $\mathbf{y}_T\sim\mathcal{N}(0, I)$, and $\mathbf{x}_0$ is fixed using the trained embedding vectors. At each sampling step $t$, we need to construct partially noised sequences as used during training, $\mathbf{z}_0$ is directly reconstructed using $f_{\theta}(\mathbf{z}_t, t)$ under the condition of $\mathbf{x}_0$ and further parameterized into $\mathbf{z}_{t-1}$ using clamping-like operation on the $\mathbf{x}_0$, it is in the same manner as training. The clamping tricks for $\mathbf{y}_t$ is the same with Diffusion-LM. 

To improve the quality of generation, we apply the widely used Minimum Bayes Risk~(MBR) decoding strategy~\citep{koehn2004statistical}. We first generate a set of candidate samples $\mathcal{S}$ from different random seeds of \textsc{DiffuSeq} and select the best output sequence that achieves the minimum expected risk under a meaningful loss function (e.g. BLEU or other cheaper metrics like precision). In practice, we use the negative BLEU score in our implementation.

% , $\hat{\mathbf{w}}^y = \argmax_{\mathbf{w}^y\in \mathcal{S}}\sum_{\mathbf{w}^{y\prime}\in \mathcal{S}}\frac{1}{|\mathcal{S}|}\text{BLEU}(\mathbf{w}^y, \mathbf{w}^{y\prime})$.

\paragraph{Connections to AR, Iter-NAR, and Fully-NAR Models.}
\label{sec:learning}
To better understand the behavior of \textsc{DiffuSeq}, we give the theoretical connection to autoregressive (AR), iterative non-autoregressive (iter-NAR), and fully non-autoregressive (fully-NAR) models. 
We argue that \model can be seen as an extension of iter-NAR model.
Detailed graphical learning discrepancies of these four cases are discussed in Appendix~\ref{sec:graph} for reference.

AR models learn $p(\mathbf{w}^y_{1:n}|\mathbf{w}^x)$ by autoregressive decomposition based on left-context:
\begin{equation}
\label{eq:ar}
    p_{\text{AR}}(\mathbf{w}^y_{1:n}|\mathbf{w}^x)=\underbrace{p(w^y_1|\mathbf{w}^x)\vphantom{\prod_{i=1,\ldots,n-1}}}_{\text{initial prediction}}\underbrace{\prod_{i=1,\ldots,n-1}p(w^y_{i+1}|\mathbf{w}^y_{1: i},\mathbf{w}^x)}_{\text{progressive left-context prediction}},
\end{equation}
while fully-NAR models~\citep{gu2017non, qian-etal-2021-glancing} learn the conditional probability given independent assumption for fast inference:
\begin{equation}
\label{eq:fully-nar}
    p_\text{fully-NAR}(\mathbf{w}^y_{1:n}|\mathbf{w}^x)=\prod_{i=1,\ldots,n}p(w^y_i|\mathbf{w}^x).
\end{equation}
To make a better analogy to AR and NAR models, we use a lossless way to formulate iterative NAR models~\citep{gu2019levenshtein, ghazvininejad-etal-2019-mask} by introducing a series of intermediate sequences $\mathbf{w}^y_{1:K-1}, \mathbf{w}^y_K=\mathbf{w}^y$ with $K$ editable iterations:
\begin{equation}
\label{eq:iter-nar}
\begin{split}
    p_\text{iter-NAR}(\mathbf{w}^y_{1:n}|\mathbf{w}^x)
    % &=\sum_{\mathbf{w}^y_1,\ldots,\mathbf{w}^y_{K-1}}{p(\mathbf{w}^y_1|\mathbf{w}^x)\prod_{k=1\ldots K-1}{p(\mathbf{w}^y_{k+1}|\mathbf{w}^y_k,\mathbf{w}^x)}} \\
    % p(\mathbf{w}^y_{1:n}|\mathbf{w}^x)&=\sum_{\mathbf{w}^y_1,\ldots,\mathbf{w}^y_{K-1}}{\underbrace{p(\mathbf{w}^y_1|\mathbf{w}^x)\vphantom{\prod_{k=1\ldots K-1}}}_{\text{unigram model}}\underbrace{\prod_{k=1\ldots K-1}{p(\mathbf{w}^y_{k+1}|\mathbf{w}^y_k,\mathbf{w}^x)}}_{\text{iterative combinatorial refinement process}}} \\
    &=\sum_{\mathbf{w}^y_1,\ldots,\mathbf{w}^y_{K-1}}{\underbrace{\prod_{i=1\ldots n}{p(w^y_{1,i}|\mathbf{w}^x)}\vphantom{\prod_{k=1..K-1}}}_{\text{initial prediction}}\underbrace{\prod_{k=1..K-1}{\prod_{i=1\ldots n}{p(w^y_{k+1,i}|\mathbf{w}^y_{k,1:n},\mathbf{w}^x)}}}_{\text{progressive full-context prediction}}}.
\end{split}
\end{equation}
Previous study~\citep{huang2022learning} shows that there is a gap called \emph{conditional total correlation} between AR Eq.~(\ref{eq:ar}) and fully-NAR Eq.~(\ref{eq:fully-nar}) learning paradigms, because of lossy decomposition of NAR models. 
However, when comparing iter-NAR Eq.~(\ref{eq:iter-nar}) with AR Eq.~(\ref{eq:ar}) models, they both can be factorized into an initial prediction term and a progressive prediction process based on different context (i.e. left-context in AR and full-context in iter-NAR), and the discrepancy pointed out by \citet{huang2022learning} is therefore closed in iter-NAR assuming sufficient steps.
By showing \model is an extension of the iter-NAR model, we offer a justification that it will not suffer from the conditional total correlation for the same reason.

A straight-forward way to formulate pure continuous diffusion models is to introduce a series of Gaussian noise-corrupted features along with diffusion steps: $\mathbf{y}_{1:T-1}, \mathbf{y}_0=\mathbf{y}, \mathbf{y}_T\sim\mathcal{N}(0, \mathbf{I})$.
\begin{equation}
\label{eq:diffusion}
p_\text{diffusion}(\mathbf{w}^y|\mathbf{w}^x)={\int_{\mathbf{y}_{T},\ldots,\mathbf{y}_0}}\underbrace{{p(\mathbf{w}^y|\mathbf{y}_{0},\mathbf{w}^x)\vphantom{\prod_{t=T,\ldots,1}}}}_{\text{final prediction}}\underbrace{\prod_{t=T,\ldots,1}{p(\mathbf{y}_{t-1}|\mathbf{y}_t,\mathbf{w}^x)}}_{\text{progressive full-context diffusion}},
\end{equation}
where $p(\mathbf{y}_{t-1}|\mathbf{y}_t, \mathbf{w}^x)$ describes the diffusion step on continuous representations $\mathbf{y}$. 
The rounding operation in \model maps the continuous vectors $\mathbf{y}$ to discrete $\mathbf{w}^y$ for each time step $t$, we in addition introduce this into Eq.~(\ref{eq:diffusion}):

\begin{align}
% \begin{split}
    p_{\model}(\mathbf{w}^y|\mathbf{w}^x)
    &={\sum_{\mathbf{w}^y_T,\ldots,\mathbf{w}^y_1}\int_{ \mathbf{y}_{T},\ldots,\mathbf{y}_0}}{p(\mathbf{w}^y|\mathbf{y}_{0},\mathbf{w}^x)\prod_{t=T,\ldots,1}{p(\mathbf{w}^y_{t}|\mathbf{y}_t,\mathbf{w}^x)p(\mathbf{y}_{t-1}|\mathbf{w}^y_t)}} \label{eq:latent-contiguous-discrete-diffusion} \\
    &={\sum_{\mathbf{w}^y_T,\ldots,\mathbf{w}^y_1}\int_{ \mathbf{y}_{T},\ldots,\mathbf{y}_0}}p(\mathbf{w}^y_T|\mathbf{y}_T,\mathbf{w}^x)\prod_{t=T-1,\ldots,0}{p(\mathbf{y}_t|\mathbf{w}^y_{t+1})p(\mathbf{w}^y_t|\mathbf{y}_t,\mathbf{w}^x)} \label{eq:latent-contiguous-discrete-nar}.
    % &=\sum_{\mathbf{y}_{T},\ldots,\mathbf{y}_0}{p(\mathbf{w}^y|\mathbf{y}_{0},\mathbf{w}^x)}\sum_{\mathbf{w}^y_{T},\ldots,\mathbf{w}^y_1}\prod_{t=T,\ldots,1}{p(\mathbf{y}_{t-1}|\mathbf{w}^y_t)p(\mathbf{w}^y_{t}|\mathbf{y}_t,\mathbf{w}^x)} \\
    % &=\sum_{\mathbf{y}_{T},\ldots,\mathbf{y}_0}{p(\mathbf{w}^y|\mathbf{y}_{0},\mathbf{w}^x)}\prod_{t=T,\ldots,1}\sum_{\mathbf{w}^y_t}{p(\mathbf{y}_{t-1}|\mathbf{w}^y_t)p(\mathbf{w}^y_{t}|\mathbf{y}_t,\mathbf{w}^x)}
% \end{split}
\end{align}
By rearranging Eq.~(\ref{eq:latent-contiguous-discrete-diffusion}) into Eq.~(\ref{eq:latent-contiguous-discrete-nar}), we can see \model can be seen as a more generalized form of iter-NAR Eq.~(\ref{eq:iter-nar}) before marginalizing out $\{\mathbf{y}_{T},\ldots,\mathbf{y}_0\}$,
% where Eq.~\ref{eq:latent-contiguous-discrete-diffusion} and Eq.~\ref{eq:latent-contiguous-discrete-nar} are equivalent with different computation order, 
despite the different initialization of $\mathbf{y}_T$\footnote{For NAR models, $\mathbf{y}_T$ is uniform copied from the source sentence or \emph{unk}'s token embedding~\citep{gu2017non}; for diffusion models, $\mathbf{y}_T$ is sampled from normal distribution $\mathcal{N}(0,\mathbf{I})$.}. A more detailed derivation is shown in Appendix~\ref{sec:nar-to-diffusion}.
% From Eq.~\ref{eq:latent-contiguous-discrete-nar} and Eq.~\ref{eq:latent-contiguous-discrete-diffusion} of \model, we derive both of iterartive NAR model~(Eq.~\ref{eq:iter-nar}) and vanilla diffusion process~(Eq.~\ref{eq:diffusion}), by marginalizing $\mathbf{y}$ and $\mathbf{w}^y$ respectively~(detailed derivation is shown in Appendix~\ref{sec:nar-to-diffusion}).
% We argue that a key point to learn diffusion process in discrete space is leveraging an step-wise alternation process of rounding $p(\mathbf{w}^y_t|\mathbf{y}_t,\mathbf{w}^x)$ and noisy embedding $p(\mathbf{y}_{t-1}|\mathbf{w}^y_t)=\mathcal{N}(\mathbf{y}_{t-1};\sqrt{1-\beta_t}\textsc{Emb}(\mathbf{w}^y_t), \beta_t \mathbf{I})$.

% For conciseness, we intuitively~\jiangtao{it is a weak claim and is conflicted to word ``theorectical'' in the other sections. it is a safe claim and is okay to me. just make sure every part of the paper are consistent.} connect AR, interative-NAR and \model by regarding them as generation process along with different steps, i.e., token-level time steps, iteration steps, diffusion steps respectively.

% It is notable that unlike AR and fully NAR models generating text all at once, iterative NAR and diffusion models features a self-corrected text generation process.
\section{Experiments}
We conduct experiments to validate the effectiveness of \model on four different tasks, against six strong AR/NAR baselines.

\begin{table}[!th]
\small
\centering
\begin{threeparttable}[b]
\caption{The overall results of different methods on different \textsc{Seq2Seq} tasks. The first group $\diamond$ of methods adopt autoregressive encoder-decoder architecture and the second group $\bullet$ is the finetuned large pre-trained language model (also in autoregressive manner) while the last group $\ddagger$ is non-autoregressive.
The best results are \textbf{bold}, and the best results without PLMs are \underline{underlined}.}
\label{tb:main}

\begin{tabular}{ll|lll|ll|c}
\toprule
Tasks               & Methods   & BLEU$\uparrow$ & R-L$\uparrow$ & Score$\uparrow$ & dist-1$\uparrow$ & selfB$\downarrow$ / div-4$\uparrow$ & Len \\
\midrule
% \midrule
\multirow{7}{*}{\makecell[l]{Open\\Domain\\Dialogue}} 
                          & GRU-attention $^{\diamond}$    & 0.0068 & 0.1054 & 0.4128 & 0.8998 & 0.8008/0.1824       & 4.46   \\
                          & Transformer-base $^{\diamond}$ & \underline{\textbf{0.0189}} & 0.1039 & 0.4781 & 0.7493 & 0.3698/0.6472    & 19.5 \\
                          \cmidrule{2-8}
                          & GPT2-base FT$^{\bullet}$& 0.0108 & \textbf{0.1508} & 0.5279 & 0.9194 & 0.0182/0.9919    & 16.8    \\
                          & GPT2-large FT $^{\bullet}$& 0.0125 & 0.1002 & \textbf{0.5293} & 0.9244 & 0.0213/0.9938    & 16.8   \\
                          & GPVAE-T5$^{\bullet}$ & 0.0110 & 0.1009 & 0.4317 & 0.5625 & 0.3560/0.5551 & 20.1 \\
                          \cmidrule{2-8}
                          & NAR-LevT $^{\ddagger}$    & 0.0158 & 0.0550 & 0.4760 & \underline{\textbf{0.9726}} & 0.7103/0.1416    & 4.11    \\
                          & \model (Ours) $^{\ddagger}$     & 0.0139 & \underline{0.1056} & \underline{0.5131} & 0.9467 & \underline{\textbf{0.0144}}/\underline{\textbf{0.9971}}    & 13.6   \\
\midrule
\midrule
\multirow{7}{*}{\makecell[l]{Question\\Generation}}       
                          & GRU-attention $^{\diamond}$    & 0.0651 & 0.2617 & 0.5222 & 0.7930 & 0.9999/0.3178       & 10.1   \\
                          & Transformer-base $^{\diamond}$ & 0.1663 & 0.3441 & \underline{0.6307} & \underline{0.9309} & 0.3265/0.7720    & 10.3 \\
                          \cmidrule{2-8}
                          & GPT2-base FT $^{\bullet}$& 0.0741 & 0.2714 & 0.6052 & 0.9602 & \textbf{0.1403}/\textbf{0.9216}    & 10.0    \\
                          & GPT2-large FT $^{\bullet}$& 0.1110 & 0.3215 & \textbf{0.6346} & \textbf{0.9670} & 0.2910/0.8062    & 9.96   \\
                          & GPVAE-T5$^{\bullet}$ & 0.1251 & 0.3390 & 0.6308 & 0.9381 & 0.3567/0.7282 & 11.4 \\
                          \cmidrule{2-8}
                          & NAR-LevT  $^{\ddagger}$  & 0.0930 & 0.2893 & 0.5491 & 0.8914 & 0.9830/0.4776    &  6.93  \\
                          & \model (Ours)$^{\ddagger}$  & \underline{\textbf{0.1731}} & \underline{\textbf{0.3665}} & 0.6123 & {0.9056} & \underline{0.2789}/\underline{0.8103}    & 11.5   \\
                          
\midrule
\midrule
\multirow{7}{*}{\makecell[l]{Text\\Simpli-\\fication}}
                          & GRU-attention $^{\diamond}$    & 0.3256 & 0.5602 & 0.7871 & 0.8883 & 0.9998/0.3313        & 18.9   \\
                          & Transformer-base $^{\diamond}$ & 0.2693 & 0.4907 & 0.7381 & 0.8886 & 0.6924/0.5095     & 18.5  \\
                          \cmidrule{2-8}
                          & GPT2-base FT $^{\bullet}$& 0.3083 & 0.5461 & 0.8021 & 0.9439 & 0.5444/0.6047    & 16.1   \\
                          & GPT2-large FT $^{\bullet}$& 0.2693 & 0.5111 & 0.7882 & 0.9464 & 0.6042/0.5876    & 15.4   \\
                          & GPVAE-T5 $^{\bullet}$     & 0.3392 & 0.5828 & \textbf{0.8166} & 0.9308 & 0.8147/0.4355    & 18.5   \\
                          \cmidrule{2-8}
                          & NAR-LevT  $^{\ddagger}$    & 0.2052 & 0.4402 & 0.7254 & \underline{\textbf{0.9715}} & 0.9907/0.3271    & 8.31   \\
                          & \model (Ours)   $^{\ddagger}$   & \underline{\textbf{0.3622}} & \underline{\textbf{0.5849}} & \underline{0.8126} & 0.9264 & \underline{\textbf{0.4642}}/\underline{\textbf{0.6604}}    & 17.7   \\
\midrule
\midrule
\multirow{7}{*}{Paraphrase}
                          & GRU-attention $^{\diamond}$    & 0.1894 & 0.5129 & 0.7763 & 0.9423 & 0.9958/0.3287       & 8.30   \\
                          & Transformer-base $^{\diamond}$ & \underline{\textbf{0.2722}} & 0.5748 & \underline{0.8381} & 0.9748 & 0.4483/0.7345    & 11.2  \\
                          \cmidrule{2-8}
                          & GPT2-base FT $^{\bullet}$ & 0.1980 & 0.5212 & 0.8246 & 0.9798 & 0.5480/0.6245    & 9.67   \\
                          & GPT2-large FT $^{\bullet}$& 0.2059 & 0.5415 & 0.8363 & \textbf{0.9819} & 0.7325/0.5020    & 9.53   \\
                          & GPVAE-T5  $^{\bullet}$  & 0.2409 & \textbf{0.5886} & \textbf{0.8466} & 0.9688 & 0.5604/0.6169    & 9.60   \\
                          \cmidrule{2-8}
                          & NAR-LevT  $^{\ddagger}$  & 0.2268 & 0.5795 & 0.8344 & 0.9790 & 0.9995/0.3329    & 8.85  \\
                          & \model (Ours)  $^{\ddagger}$  & 0.2413 & \underline{0.5880} & 0.8365 & \underline{0.9807} & \underline{\textbf{0.2732}}/\underline{\textbf{0.8641}}    & 11.2  \\
\bottomrule
\end{tabular}
% \begin{tablenotes}
% \end{tablenotes}
\end{threeparttable}
\end{table}
% \paragraph{Qualitative Analysis}

\subsection{Experimental Setup}

\paragraph{Tasks and Datasets.}
\textsc{Seq2Seq} generation covers a wide range of tasks, among which we choose four typical and popular tasks.
\textbf{Open domain dialogue} requires models to generate informative responses given a dialogue context. We use Commonsense Conversation Dataset~\citep{zhou2018commonsense}, which is extracted from Reddit single-round dialogs, with over 3 million conversational pairs.
\textbf{Question generation}(QG) aims to generate questions given a context as input. To obtain sufficient training samples, we use the dataset Quasar-T~\citep{dhingra2017quasar} preprocessed by~\citet{lin2018denoising}, and then generate document-question pairs to obtain 119K training samples (details in Appendix~\ref{sec:appendix-qg}).
\textbf{Text simplification} aims to revise the complex text into sequences with simplified grammar and word choice. \citet{jiang2020neural} constructs a corpus consisting of 677K complex-simple sentences with revision alignment. 
% In some training data the same complex could correspond to multiple references.
\textbf{Paraphrase} task generates an alternative surface form in the same language expressing the same semantic content. We adopt widely used QQP
\footnote{\url{https://www.kaggle.com/c/quora-question-pairs}}
 sourced from the community question answering forum Quora, with 147K positive pairs.
% \subsection{Experiments Setup}
\paragraph{Baselines.}
We consider three groups of models as baselines, covering both AR and NAR architectures. The first group of methods adopts encoder-decoder architecture~\citep{cho2014learning} which is well-studied for \textsc{Seq2Seq} tasks, and we conduct experiments on two popular models: GRU with attention and Transformer~\citep{vaswani2017attention}. The second group is the finetuned large pre-trained language model (PLM), among which GPT2~\citep{radford2019language} has demonstrated great success in almost all \textsc{Seq2Seq} tasks. We further compare to GPVAE~\citep{du2022diverse}, which augments a pre-trained T5~\citep{2020t5} with VAE to improve the generation diversity. 
For the last group of baselines, we consider LevT~\citep{gu2019levenshtein}, a widely used, strong iterative NAR model. 
All baselines are trained following instructions in their papers, and details can be found in Appendix~\ref{sec:appendix-basline}.

\paragraph{Evaluation.}
We evaluate the generated sequences from two aspects: quality and diversity.
To evaluate the quality, we use the standard metric BLEU~\citep{papineni-etal-2002-bleu} and ROUGE~\citep{lin2004rouge} score. 
% which compute string overlappings with respect to ground references.
% The BLEU score is sentence-level smoothed from BLEU-1 to 4, and ROUGE-L score is longest common subsequence based statistics. 
% The n-gram based metrics may fail to capture the semantic meaning of sentences, 
Since string-similarity-based metrics can be unsatisfactory for open-ended generation, we also report BERTScore~\citep{zhang2019bertscore} that assesses the semantic similarity between generated sentences and references. Details are in Appendix~\ref{sec:appendix-metrcis}.
Higher scores of BLEU, ROUGE and BERTScore reflect better performance.
As for diversity, we use distinct unigram~(dist-1) to measure intra-diversity within each generated sentence, where the lower dist-1 indicates that the generated sentence contains more repeated words. 
For sentence-level diversity evaluation, we consider sentence-level self-BLEU~\citep{zhu2018texygen} to measure the n-gram overlap between the set of outputs w.r.t one source sentence, and we additionally use diverse 4-gram (div-4)~\citep{deshpande2019fast} to measure the ratio of distinct 4-grams in the set of outputs per source sentence. 
The lower self-BLEU and higher div-4 suggest higher diversity of generation. For each method including \model, we generate $3$ samples for each source sentence to compute the diversity metrics. 

\paragraph{Implementation Details.}
Our \model is based on the $12$ layers of Transformer with $12$ attention heads, where the time step embedding is plugged akin to the position embedding. The maximum sequence length is $128$, with embedding dimension $d=128$, diffusion steps $T = 2,000$ and a square-root noise schedule. To reduce the out-of-vocabulary generation, we apply Byte Pair Encoding~\citep{sennrich-etal-2016-neural} to construct the vocabulary. After conducting the diversity beam search (DBS)~\citep{vijayakumar2016diverse} for the Transformer-base model and GPT model, we find that DBS does not always promote diversity over temperature sampling and therefore we list the best diversity results.
{We compute the accuracy metrics of \model using MBR with the size of candidate samples $|\mathcal{S}|=10$. The experiment is deployed on NVIDIA A100 Tensor Core GPUs, and we use 4 GPUs on training and single GPU on sampling.}
\subsection{Main Results}
\label{sec:result}
% \paragraph{Automatic Evaluation}
% The overall results on four different \textsc{Seq2Seq} tasks is shown in~\tabref{tb:main}. We can take several observations. \model achieve comparable quality score with PLMs while maintains high diversity.
As shown in~\tabref{tb:main}, we conclude that \model achieves comparable or even higher generation quality compared with strong baselines. 
At the same time, \model consistently demonstrates its superiority in generating diverse outputs given the same input sequence.

As we can see from ~\tabref{tb:main}, \model wins competitions over at least one quality metric against 6 baselines $\times$ 4 tasks. Although NAR models such as LevT can also outperform AR baselines sometimes, they still lag well behind \model by large margins (i.e., relative improvements over 50$\%$ for BLEU in QG task and R-L in Dialogue task). Even compared with pre-trained then finetuned GPT2 models, \model still delivers superior performance than the base variant, and is comparable with the large variant, which has $8.2$ times more parameters than \model. These empirical results amply support our findings in~\Secref{sec:learning}, where we theoretically analyze the potential of diffusion models in modeling text sequences compared with AR models given sufficient diffusion steps.

\model, as a member of the deep generative model family, also exhibit the capacity to generate highly diverse sequences. As suggested by self-BLEU~(lower is better) and div-4~(higher is better), in almost all cases, \model significantly outperforms 4 AR baselines in terms of sentence-level diversity~(i.e., producing diverse outputs given the same input). For diversity in word choice within one sentence, we consider dist-1: a higher dist-1 indicates less repetition within a sentence. As we can see from~\tabref{tb:main}, \model has less repetition compared with encoder-decoder methods, but still fall behind the pre-trained GPT2 models (the same situation with BERTScore). These results suggest there is still room for improvement (e.g., use pre-training techniques) in diffusion models' token-level choice. Different from NAR-LevT, \model does not rely on an extra length prediction module but automatically decides by the padding token instead and is able to generate longer output sentences, indicated by the last column for average generation length.

\begin{figure}[b]
\centering
\begin{minipage}[t]{0.46\textwidth}
% \raggedleft
% \flushleft
\centering
\includegraphics[width=6.7cm]{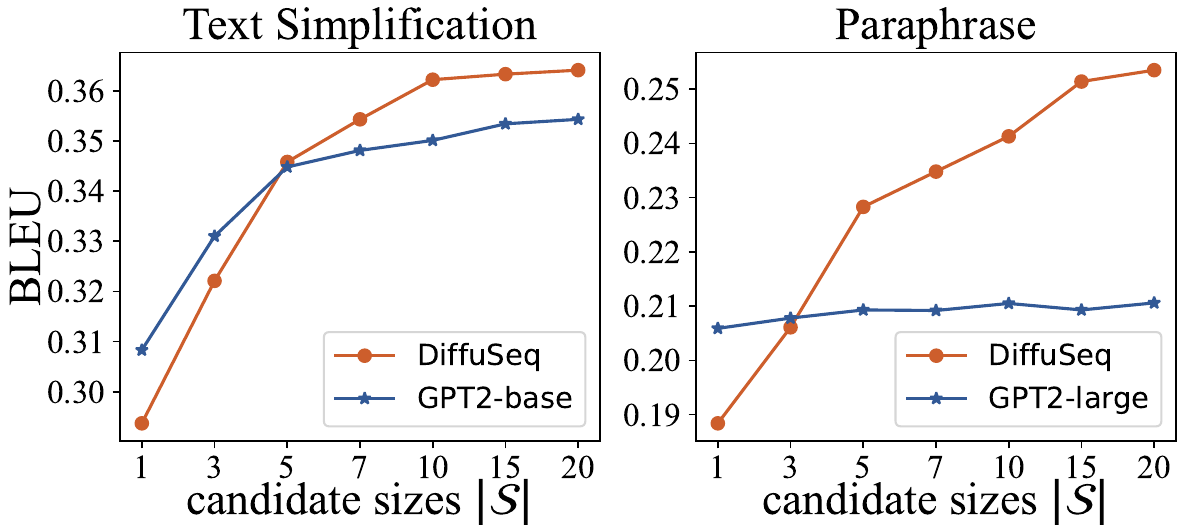}
\caption{The increase of BLEU score with different candidate sizes $|\mathcal{S}|$.}
\label{fig:div_rank}
\end{minipage}
\hspace{0.5cm}
\begin{minipage}[t]{0.46\textwidth}
\centering
% \flushright
% \raggedright
\makebox[\textwidth][r]{\includegraphics[width=6.7cm]{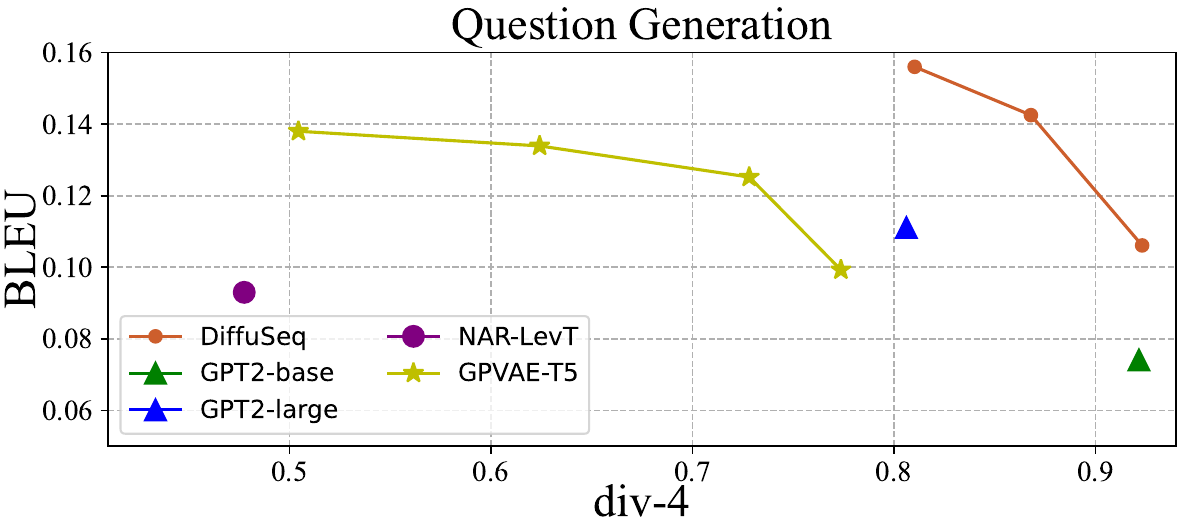}}
\caption{Trade-off between quality and diversity (details in Appendix~\ref{sec:appendix-trade-off}).}
\label{fig:trade_off_qual_div}
\end{minipage}
\end{figure}

In~\tabref{tb:case}, we provide examples to showcase \textsc{DiffuSeq}'s ability to generate diverse samples. More examples can be found in Appendix~\ref{sec:appendix-case}.

% As for the dist-1 score, \model is not always the highest but compared to vanilla encoder-decoder methods, it is able to generate less repeat words. Besides, \model tends to generate longer sequences, and the generation length is automatically decided by \model. This also differs from the NAR, which relies on an extra length prediction module.

\begin{table}[tb]
\small
\centering
\begin{threeparttable}[b]
\caption{Sample outputs in QQP test set, conditioned on the same $\textbf{x}$.}
\label{tb:case}

\begin{tabular}{l|l|l}
\toprule
\multicolumn{3}{l}{\textit{\textbf{Original sentence}: How do I make friends.\quad \quad \textbf{Paraphrase reference}: How to make friends ?}}\\
% \multicolumn{2}{l}{\textit{Paraphrase reference: How to make friends ?}}\\
\midrule
% \midrule

\textbf{GPT2-large finetune} & \textbf{GPVAE-T5} & \textbf{\model} \\
 {How can I make friends?} &  {How can I make friends?} &  {How can I make friends better?} \\ 
 {How can I make friends?} &  {How do I make friends?}  &  {How can I make friends?}\\ 
 {How can I make friends?} & {How can I make friends?} &  {How do you make friends?} \\
 {How can I make friends?} &  {How can I make friends?} &  {What is the best way to make friends?}\\
\makecell[l]{How do I make friends and\\ keep them?} & \makecell[l]{What's the best way to make \\friends and make make friends?}  &  \makecell[l]{How can I make friends and more \\something?}\\
\bottomrule
\end{tabular}
\end{threeparttable}
\end{table}

\subsection{Analysis}

% Given the superiority of \model in generating highly-diverse sequences, it is natural to ask: How does generation diversity offered by \model relates to quality and what are the involved trade-offs?
% \jiangtao{not all of the analysis is on diversity.}

% We conduct a series of analysis, including the relation between diversity offered by \model and generation quality as well as generation process, and other settings of \model.

We conduct a series of analysis to investigate the effectiveness of different aspects in \model.

\paragraph{Diversity Ensures Quality.} 
Generating high-quality texts with high diversity is an important requirement for many text generation applications and the trade-off between quality and diversity is always a critical concern in open-ended NLG tasks~\citep{zhang-etal-2021-trading}. Different from AR models relying on the decoding strategy like temperature and nucleus sampling~\citep{holtzman2019curious} and VAE models sampling latent variable from Gaussian Prior, the natural advantage of \model is to generate different sentences along with a series of random Gaussian noise.
In~\Figref{fig:trade_off_qual_div}, we elucidate that \model have better trade-off between generation quality (BLEU) and sentence-level diversity (div-4).
Here we further demonstrate that the high diversity provided by \model can be turned into better quality. 

MBR is a common strategy to improve generation quality by aggregating and ranking candidate sequences, and we find that the upper bound of MBR is decided by a diversified candidate set. To valid this, we simultaneously apply MBR on both \model and GPT2 with various candidate sizes $|\mathcal{S}|$. The results are shown in~\Figref{fig:div_rank}. As we can see, \model lags behind GPT2 without using MBR ($|\mathcal{S}|=1$) or with a small candidate set ($|\mathcal{S}|=3$). However, as $|\mathcal{S}|$ increases, \model starts to outperform GPT2 by an increasing margin. The reason is that autoregressive models like GPT2 tend to generate highly similar candidates (as discussed in \S~\ref{sec:result}), which impedes the effectiveness of MBR. As $|\mathcal{S}|$ increases to $20$, \model still shows better 
rising trends than GPT2. Our findings also stress the importance of better ranking methods in diffusion research.

\begin{figure}[tb]
\centering
\begin{minipage}[t]{0.55\textwidth}
\centering
\includegraphics[height=3.2cm]{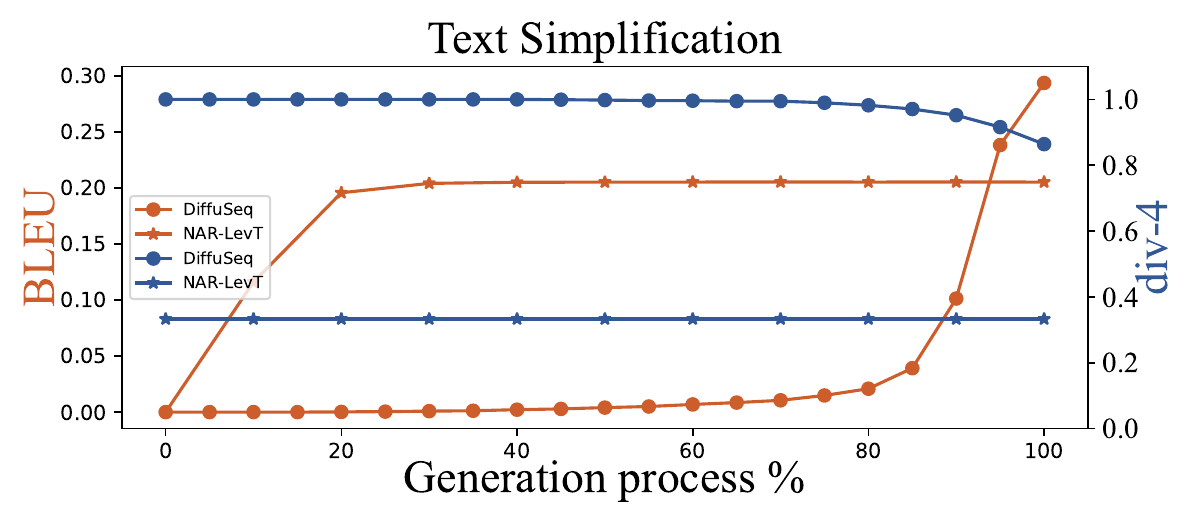}
\caption{The curve of BLEU/div-4 score along with generation process (percentage of steps).}
\label{fig:step_bleu}
\end{minipage}
\hspace{0.02\textwidth}
\begin{minipage}[t]{0.38\textwidth}
\centering
\makebox[\textwidth][c]{\includegraphics[height=3.2cm]{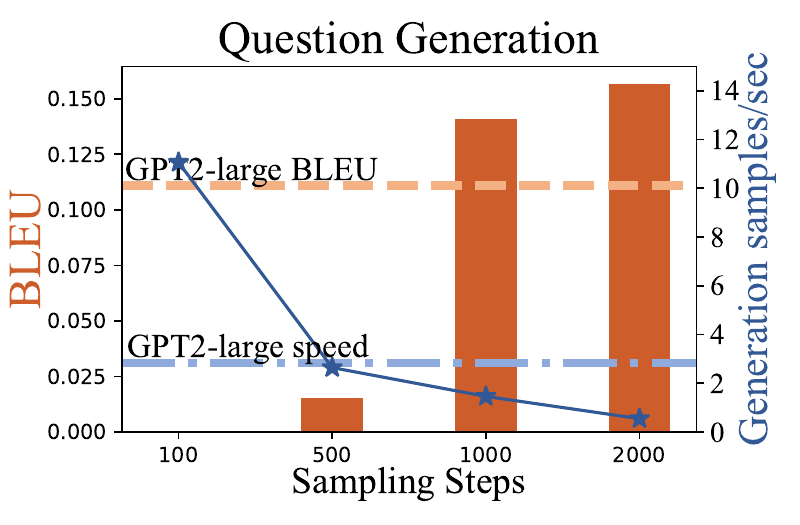}}
\caption{ {The BLEU and inference speed of \model and GPT2-large.}}
\label{fig:speed}
\end{minipage}
\end{figure}

\paragraph{Step-wise Analysis against Iterative NAR.} 
% \jiangtao{
Given the underlying theoretical connection between iterative NAR and \model discussed in ~\Secref{sec:learning},
we empirically investigate the behavior of LevT and \model by analyzing their step-wise quality (i.e. BLEU) and diversity (i.e. div-4) curves.
As is suggested in~\Figref{fig:step_bleu}, LevT grows fiercely in quality at the very beginning of generation, and quickly slows down in the successive refinement process.
But \model behaves differently, with BLEU score growing slowly at first, increasing rapidly as the diffusion process progresses and finally surpassing LevT.
It is also observed that the diversity of both LevT and \model is determined at the very early stage regardless of future refinement or diffusion, where \model consistently outperforms LevT on diversity at any stage of generation.
% }
% As we discussed in~\Secref{sec:learning}, \model are highly related with iterative-NAR, and the difference is that one is along with the diffusion steps $t$ and another is along with the iterative steps $k$ when generating. We demonstrate the changing of BLEU and Self-BLEU step-wisely in~\Figref{fig:step_bleu}. Iterative-NAR LevT achieves high BLEU at the first few steps, while for \model the BLEU score grows steadily at the first half of steps and then increase rapidly at the last half of steps. The self-BLEU score behaves similarly. \mukai{trend of div-4 score is opposed to BLUE}
We conjecture that \model explores more possible results at the first half of generation process, and soon converges to several potential candidates when it is closed to the end of steps. In this case, \model shows its capacity to take both generation quality and diversity into consideration, and this is the capacity that iterative-NAR and even AR models can not obtain, due to the different learning paradigms.

% \subsection{Ablation Study}

\paragraph{Inference Speed.} The slow sampling speed is one of the major concerns about diffusion models. Here we fix the number of diffusion steps during training for \model while shrinking the inference steps following DDIM~\citep{song2020denoising}. As we can see from~\Figref{fig:speed}, when reducing the inference to 1,000 diffusion steps  {on single GPU}, \model achieves a higher BLEU score than GPT2-large yet registers a  {closer} inference speed to GPT2-large. 
% With recent advances in improving inference speed of diffusion models (citation), we are optimistic that diffusion models have the potential to outperform popular PLMs in both generation quality and inference speed. \zy{delete last sent probably}

\paragraph{Effectiveness of Joint Training.} 
% Note that text-to-image diffusion models~\citep{nichol2021glide, ramesh2022hierarchical} also concatenate two representations\zy{not sure whether this is correct or not}(image and text) in their designs to perform conditional image generation.   except that the text representation is frozen during training.

In \model, the representations of $\textbf{w}^x$ and $\textbf{w}^y$ are jointly trained using the same embedding function $\textsc{Emb}(\cdot)$ (stated in~\Secref{sec:learning}). To validate the effectiveness of this joint training strategy, we compared it with the training strategy commonly used in text-to-image diffusion models~\citep{nichol2021glide, ramesh2022hierarchical}. In particular, we decouple the training of $\textsc{Emb}(\textbf{w}^x)$ and $\textsc{Emb}(\textbf{w}^y)$ by replacing $\textsc{Emb}(\textbf{w}^x)$ with representations extracted from a pre-trained BERT-tiny model~\citep{DBLP:journals/corr/abs-1908-08962}. From~\tabref{tb:fixx}, we find that the decoupled training strategy results in poor performance. 
% Our results indicate that directly applying previous success in vision domain 

% we use the fixed $\textsc{Emb}(\textbf{w}^x)$ from pre-trained BERT-tiny while keeping the jointly learning of diffusion model's parameters and word embeddings of $\textbf{w}^y$, which is in the similar setting with text-to-image diffusion models~\citep{nichol2021glide, ramesh2022hierarchical}. We find that fixed embeddings for input sequence $\textbf{w}^x$ are suboptimal for \model compared to jointly training.

\begin{table}[t]
\centering
\begin{threeparttable}[b]
\caption{Results with or without joint training for Question Generation task.}
\label{tb:fixx}
\begin{tabular}{l|llll}
\toprule
Setting & BLEU$\uparrow$ & R-L$\uparrow$ & Score$\uparrow$ & selfB$\downarrow$ / div-4$\uparrow$ \\
\midrule
\model (w/o reranking) & 0.1567 & 0.3484 & 0.5947 & 0.2789/0.8103 \\
Fix $\textsc{Emb}(\textbf{w}^x)$ as pre-trained & 0.0110 & 0.0687 & 0.3769 & 0.0174/0.9376\\

\bottomrule
\end{tabular}
\end{threeparttable}
\end{table}
\section{Related Work}
\label{sec:related}
\paragraph{Diffusion Models for Text Modeling.}
Text-to-Image generation using diffusion models has developed many potential applications. Models such as Imagen~\citep{saharia2022photorealistic} and DALL-E~\citep{ramesh2022hierarchical} are usually two-staged relying on the pre-trained models, requiring the alignment between the embedding vectors from two sources. GLIDE~\citep{nichol2021glide} explores diffusion model with classifier-free~\citep{ho2022classifier} guidance by setting guidance scale during training. The target space of these models is not discrete text space but stable vectors of pixel values. There are other works of diffusion on text generation, but they stick to the original encoder-decoder architecture and the diffusion process is interspersed on the decoder~\citep{savinov2021step}, or the latent space~\citep{yu2022latent}.

For text generation using the diffusion models, \citet{hoogeboom2021argmax} introduce the multinomial diffusion for character-level text generation, the forward categorical noise is applied through the Markov transition matrix. \citet{austin2021structured} generalize discrete text diffusion models by introducing the absorbing state (\texttt{[MASK]}). However, discrete diffusion models may suffer from the scaling of the one-hot row vectors, and they only generate text samples unconditionally in discrete space. Diffusion-LM~\citep{li2022diffusion} and Analog Bits~\citep{chen2022analog} propose a new language model diffused on the continuous latent representations, with different mapping functions that connect the discrete and continuous space of texts. Compared with our work, we focus on the \textsc{Seq2Seq} diffusion models for text generation in the continuous space and our work is the first to explore this setting to the best of our knowledge.

\paragraph{Diffusion Models for Conditional Generation.}
Related to conditional-VAE~\citep{zhao2017learning}, we can consider the latent encoded input $\mathbf{x}$ as a condition. Diffusion-LM~\citep{li2022diffusion} adopts the plug-and-play approaches~\citep{sumanth2020plug} to compose fine-grained constraints on the generated sentences, but it fails to condition on the whole source sentence in \textsc{Seq2Seq} tasks. Noted that this controllable generation method is orthogonal to our \textsc{DiffSeq}, in other words, we can further add classifier-guided constraints on the \textsc{Seq2Seq} output to further control the text generation. 
There are other conditional diffusion models on the time series prediction like CSDI~\citep{tashiro2021csdi} or audio generation like WaveGrad~\citep{chen2020wavegrad}, but their class conditions are usually attributes that are easy to model, while the contextual texts as conditions are much more complex.
\section{Conclusions}
% 1. Further, to unify the conditional generation as well as the unconditional free text generation, we can adopt dropout training, for condition sentence with 10\%\~20\% drop, use special token instead.

We propose \model to tackle \textsc{Seq2Seq} tasks in a diffusion way, which contains the strong potential to achieve better generation quality and diversity trade-off. 
% In other words, \model is capable of producing highly diverse sentences without sacrificing much of the quality.
The capability enables favorable characteristics of \model to further enhance the quality of final results, by leveraging a minimum Bayes risk decoding algorithm.
% The diversity of \model can boost its own quality using MBR and can also bolster many downstream tasks that requires multiple references.  
Besides, we theoretically connect the AR and NAR models to \textsc{DiffuSeq}, and show that \model is a powerful extension of iterative-NAR model.
The empirical results demonstrate that \model is also a powerful model for text generation, matching or even surpassing competitive AR, iterative NAR, and large-scale pre-trained models on quality and diversity.
% Besides, AR models suffer from the generation quality and speed when generated sequences become very long, and under the new generation paradigm of \model we have the promise to bridge this flaw. (future: pre-trained diffusion models)
Given the limited progress of current diffusion models on text generation, our study addresses promising achievements by such a new sequence-to-sequence learning paradigm.


\section*{Acknowledgments}
We would like to thank the anonymous reviewers and other peers for their valuable advice, and we also acknowledge Chenxin An's efforts to update the generation results for the Transformer-base model on QG and Paraphrasing tasks. This work is partially supported by the Shanghai Committee of Science and Technology (Grant No. 21DZ1100100) and the joint research scheme of the National Natural Science Foundation of China (NSFC) and the Research Grants Council (RGC) under grant number N\_HKU714/21.

\bibliography{iclr2023_conference}
\bibliographystyle{iclr2023_conference}

\newpage
\appendix
\section{Objective Derivations Of DiffuSeq}
\label{sec:appendix-obj}
The diffusion model is well-known as its ability to achieve the trade-off between flexibility and tractability of the models' probability distributions, compared with GAN, VAE and Flow-based models. Following~\citet{ho2020denoising,nichol2021improved,song2020denoising}, we systematically define the forward noising process and reverse denoising process on latent continuous space $\mathbf{z}$. 

The \textit{forward} noising is to perturb the structure of data $\mathbf{z}_0$. $\mathbf{z}_0$ is finally changed into the partial Gaussian noise with $\mathbf{y}_T\sim \mathcal{N}(0, \mathbf{I})$ through $T$-step forward random disturbance 
\begin{equation}
    q(\mathbf{z}_{t} \vert \mathbf{z}_{t-1}) = \mathcal{N}(\mathbf{z}_{t};\sqrt{1-\beta_t}\mathbf{z}_{t-1}, {\beta}_t \mathbf{I}),
\end{equation}
with $t = 1, 2,...,T$ and $\{\beta_t \in (0,1)\}_{t=1}^T$ are the variance schedule. Let $\alpha_t=1-\beta_t$ and $\bar{\alpha}_t = \prod_{i=1}^t \alpha_i$, we have:

\begin{equation}
\begin{aligned}
\label{eq:zt}
    \mathbf{z}_t = &\sqrt{\alpha_t} \mathbf{z}_{t-1}+\sqrt{1-\alpha_t}\epsilon_{t-1}=\sqrt{\alpha_t\alpha_{t-1}} \mathbf{z}_{t-2}+\sqrt{1-\alpha_t\alpha_{t-1}}\bar{\epsilon}_{t-2}\\
    =&...=\sqrt{\bar{\alpha_t}}\mathbf{z}_0+\sqrt{1-\bar{\alpha}_t}\epsilon,
\end{aligned}
\end{equation}

where $\epsilon$ stands for Gaussian noises. In the end, $q(\mathbf{z}_t \vert \mathbf{z}_0) = \mathcal{N}(\mathbf{z}_t; \sqrt{\bar{\alpha}_t} \mathbf{z}_0, (1 - \bar{\alpha}_t)\mathbf{I})$. We use a sqrt noise schedule in Diffusion-LM~\citep{li2022diffusion}, that is, $\bar{\alpha}_t=1-\sqrt{t/T+s}$ with $s$ as a small constant at the start of noise level.
The \textit{reverse} process then denoises $\mathbf{z}_t$, aiming to recover original $\mathbf{z}_0$, and is defined as:
\begin{equation}
    p_{\theta}(\mathbf{z}_{0:T}):=p(\mathbf{z}_T)\prod_{t=1}^Tp_{\theta}(\mathbf{z}_{t-1}|\mathbf{z}_t), \quad
    p_{\theta}(\mathbf{z}_{t-1}|\mathbf{z}_t)=\mathcal{N}(\mathbf{z}_{t-1};\mu_{\theta}(\mathbf{z}_t, t), \sigma_{\theta}(\mathbf{z}_t, t)).
\end{equation}
The learning of $p_{\theta}$ is based on our diffusion model $\textsc{DiffuSeq}$: $f_{\theta}(\mathbf{z}_t, t)$, where the $\mu_{\theta}(\cdot)$ and $\sigma_{\theta}(\cdot)$ is the predicted parameterization of the mean and standard variation of $ q(\mathbf{z}_t|\mathbf{z}_{t-1})$ in forward process. Using Bayes' rule:

\begin{equation}
\begin{aligned}
    q(\mathbf{z}_{t-1} \vert \mathbf{z}_t, \mathbf{z}_0) 
    &= q(\mathbf{z}_t \vert \mathbf{z}_{t-1}, \mathbf{z}_0) \frac{ q(\mathbf{z}_{t-1} \vert \mathbf{z}_0) }{ q(\mathbf{z}_t \vert \mathbf{z}_0) }
\end{aligned}
\end{equation}

Substitute Eq.~(\ref{eq:zt}) to it and we can get the parameterized mean of $q(\mathbf{z}_{t-1} \vert \mathbf{z}_t, \mathbf{z}_0)$: 
\begin{equation}
\label{eq:ut}
    \mu_t(\mathbf{z}_t,\mathbf{z}_0)=\frac{\sqrt{\alpha_t}(1-\bar{\alpha}_{t-1})}{1-\bar{\alpha}_t}\mathbf{z}_t+\frac{\sqrt{\bar{\alpha}_{t-1}}\beta_t}{1-\bar{\alpha}_t}\mathbf{z}_0,
\end{equation}
and for brevity, we short the coefficient of $\mathbf{z}_t$ and $\mathbf{z}_0$ as $\mathcal{U}$ and $\mathcal{E}$ respectively.

We can use the variational lower bound to optimize the negative log-likelihood $\mathbb{E}[-\log p_{\theta}(\mathbf{x}_0) ]\leq \mathcal{L}_\text{VLB}$. The objective can be further rewritten to be a combination of several KL-divergence and entropy terms following~\citet{sohl2015deep}.

\begin{equation}
\begin{aligned}
\mathcal{L}_\text{VLB} =
\mathcal{L}_T + \mathcal{L}_{T-1} + \dots + \mathcal{L}_0
&=\mathbb{E}_{ q(\mathbf{z}_{1:T}|\mathbf{z}_0)}
\Bigg[
\log\frac{ q(\mathbf{z}_T|\mathbf{z}_0)}{p_{\theta}(\mathbf{z}_T)}  + \sum_{t=2}^T \log{\frac{ q(\mathbf{z}_{t-1}|\mathbf{z}_0,\mathbf{z}_t)}{p_{\theta}(\mathbf{z}_{t-1}|\mathbf{z}_t)}} \\
& +  {\log\frac{q_{\phi}(\mathbf{z}_0|\mathbf{w}^{x\oplus y})}{p_{\theta}(\mathbf{z}_0|\mathbf{z}_1)}}-\log p_{\theta}(\mathbf{w}^{x\oplus y}|\mathbf{z}_0)\vphantom{\log{\frac{ q(\mathbf{z}_{t-1}|\mathbf{z}_0,\mathbf{z}_t)}{p_{\theta}(\mathbf{z}_{t-1}|\mathbf{z}_t)}}}
\Bigg].\\
\end{aligned}
\end{equation}
For $1 \leq t \leq T-1$, we compute the parameterization of $\mathcal{L}_t$ by substituting Eq.~(\ref{eq:ut}) to minimize the difference from $\mu_t$ and $\mu_{\theta}$ following~\citet{ho2020denoising}:
\begin{equation}
\begin{aligned}
\mathcal{L}_t&=\mathbb{E}_{\mathbf{z}_0}\left[\log{\frac{ q(\mathbf{z}_{t}|\mathbf{z}_0,\mathbf{z}_{t+1})}{p_{\theta}(\mathbf{z}_{t}|\mathbf{z}_{t+1})}}\right]= \mathbb{E}_{\mathbf{z}_0}\left[\frac{1}{\mathcal{C}}|| \mu_t(\mathbf{z}_t,\mathbf{z}_0)-\mu_{\theta}(\mathbf{z}_t, t)||^2 \right]\\ & =\mathbb{E}_{\mathbf{z}_0}\left[\frac{1}{\mathcal{C}}||\;\mathcal{U}\mathbf{z}_t+\mathcal{E}\mathbf{z}_0-(\mathcal{U}\mathbf{z}_t + \mathcal{E}f_{\theta}(\mathbf{z}_t, t))||^2\right]=\frac{\mathcal{E}}{\mathcal{C}}\mathbb{E}_{\mathbf{z}_0}[||\mathbf{z}_0-f_{\theta}(\mathbf{z}_t, t)||^2],  \\
\end{aligned}
\end{equation}
where $\mathcal{C}=2||\sigma_{\theta}||^2$ is a loss independent constant. Then the optimization of training loss $\min_{\theta}\;\mathcal{L}_{\text{VLB}}$ can be further simplified as:
% $$\min_{\theta}\; \mathcal{L}_{\text{VLB}} = \min_{\theta}\left[
% % ||\mu(\mathbf{z}_T)||^2+
% \sum_{t=2}^T||\mathbf{z}_0-f_{\theta}(\mathbf{z}_t, t)||^2 + ||\textsc{Emb}(\mathbf{w}^{x\oplus y})-f_{\theta}(\mathbf{z}_1, 1)||^2-\log p_{\theta}(\mathbf{w}^{x\oplus y}|\mathbf{z}_0)\right] $$
\begin{equation}
% \label{eq:loss}
\begin{aligned}
&\min_{\theta}\left[
||\mu(\mathbf{z}_T)||^2+
\sum_{t=2}^T||\mathbf{z}_0-f_{\theta}(\mathbf{z}_t, t)||^2 + ||\textsc{Emb}(\mathbf{w}^{x\oplus y})-f_{\theta}(\mathbf{z}_1, 1)||^2-\log p_{\theta}(\mathbf{w}^{x\oplus y}|\mathbf{z}_0)\right]\\ 
\rightarrow& \min_{\theta}\left[
\sum_{t=2}^T||\mathbf{z}_0-f_{\theta}(\mathbf{z}_t, t)||^2 + ||\textsc{Emb}(\mathbf{w}^{x\oplus y})-f_{\theta}(\mathbf{z}_1, 1)|| {^2}-\log p_{\theta}(\mathbf{w}^{x\oplus y}|\mathbf{z}_0)\right]\\
\rightarrow& \min_{\theta} \left[\sum_{t=2}^T||\mathbf{y}_0-\tilde f_{\theta}(\mathbf{z}_t, t)||^2 + ||\textsc{Emb}(\mathbf{w}^y)-\tilde f_{\theta}(\mathbf{z}_1, 1)|| {^2} + \mathcal{R}(||\mathbf{z}_0||^2)\right].
\end{aligned}
\end{equation}

\section{Graphical Models of AR, Fully NAR, iterative NAR and \textsc{DiffuSeq} models}
We start from the conditional sequence generation problem, which aims to learn a conditional probability $p(\mathbf{w}^y_{1:n}|\mathbf{w}^x)$ with $\mathbf{w}^x$ and $\mathbf{w}^y$.
AR models learn $p(\mathbf{w}^y_{1:n}|\mathbf{w}^x)$ by autoregressive decomposition based on left-context:
\begin{equation}
    p_{\text{AR}}(\mathbf{w}^y_{1:n}|\mathbf{w}^x)=\underbrace{p(w^y_1|\mathbf{w}^x)\vphantom{\prod_{i=1,\ldots,n-1}}}_{\text{initial prediction}}\underbrace{\prod_{i=1,\ldots,n-1}p(w^y_{i+1}|\mathbf{w}^y_{1: i},\mathbf{w}^x)}_{\text{progressive left-context prediction}},
\end{equation}
consisting of an initial prediction and an autoregressive left-context prediction process, while fully-NAR models~\citep{gu2017non, qian-etal-2021-glancing} learn the conditional probability given independent assumption for fast inference:
\begin{equation}
    p_\text{fully-NAR}(\mathbf{w}^y_{1:n}|\mathbf{w}^x)=\prod_{i=1,\ldots,n}p(w^y_i|\mathbf{w}^x).
\end{equation}
To make a better analogy to AR and NAR models, we use a lossless way to formulate iterative NAR models~\citep{gu2019levenshtein, ghazvininejad-etal-2019-mask} by introducing a series of intermediate sequences $\mathbf{w}^y_{1:K-1},\mathbf{w}^y_K=\mathbf{w}^y$ as:

\begin{equation}
\label{eq:iter-nar2}
\begin{split}
    p_\text{iter-NAR}(\mathbf{w}^y_{1:n}|\mathbf{w}^x)&=\sum_{\mathbf{w}^y_1,\ldots,\mathbf{w}^y_{K-1}}{p(\mathbf{w}^y_1|\mathbf{w}^x)\prod_{k=1\ldots K-1}{p(\mathbf{w}^y_{k+1}|\mathbf{w}^y_k,\mathbf{w}^x)}} \\
    &=\sum_{\mathbf{w}^y_1,\ldots,\mathbf{w}^y_{K-1}}{p(\mathbf{w}^y_1|\mathbf{w}^x)\vphantom{\prod_{k=1\ldots K-1}}\prod_{k=1\ldots K-1}{p(\mathbf{w}^y_{k+1}|\mathbf{w}^y_k,\mathbf{w}^x)}} \\
    &=\sum_{\mathbf{w}^y_1,\ldots,\mathbf{w}^y_{K-1}}{\underbrace{\prod_{i=1\ldots n}{p(w^y_{1,i}|\mathbf{w}^x)}\vphantom{\prod_{k=1..K-1}}}_{\text{initial prediction}}\underbrace{\prod_{k=1..K-1}{\prod_{i=1\ldots n}{p(w^y_{k+1,i}|\mathbf{w}^y_{k,1:n},\mathbf{w}^x)}}}_{\text{progressive full-context prediction}}}
\end{split}
\end{equation}

Previous study~\citep{huang2022learning} shows that there is a gap called \emph{conditional total correlation} between AR and fully-NAR learning paradigms, because of the lossy decomposition of NAR models. This gap is mainly responsible for the performance drop from AR to NAR models.  
However, when comparing iter-NAR, Eq.~(\ref{eq:iter-nar2}), with AR models, they both can be factorized into an initial prediction term and a progressive prediction process based on different context (i.e. left-context in AR and full-context in iter-NAR). The discrepancy as pointed out by \citet{huang2022learning} is therefore closed in iter-NAR assuming sufficient steps. By showing \model is an extension of the iter-NAR model, we offer a justification that it will not suffer from the conditional total correlation for the same reason.

% The iterative NAR in Eq.~\ref{eq:iter-nar} consists of a unigram prediction and an iterative combinatorial refinement process.
% However, we argue that the analysis only suits fully-NAR models, but not iterative models, such as iterative-NAR and \model, which inherently violate the independent assumption.
% The unigram model learns to generate initial tokens independently, and is believed as powerful as $p(w^y_1|\mathbf{w}^x)$ in AR model; the iterative refinement process learns to optimize a pre-decoded sequences combinatorially, with richer contexts than AR models.
% Thus it can be inferred that iterative NAR models do not suffer from \emph{conditional total correlation}.
A straight-forward way to formulate naive diffusion models is to introduce a series of Gaussian noise-corrupted features $\mathbf{y}_{1:T-1}, \mathbf{y}_0=\mathbf{y}, \mathbf{y}_T\sim\mathcal{N}(0, \mathbf{I})$ on continuous space as:
\begin{align}
\label{eq:diffusion2}
p_\text{diffusion}(\mathbf{w}^y|\mathbf{w}^x)
&= {\int}_{\mathbf{y}_{T},\ldots,\mathbf{y}_0}\underbrace{{p(\mathbf{w}^y|\mathbf{y}_{0},\mathbf{w}^x)\vphantom{\prod_{t=T,\ldots,1}}}}_{\text{final-step prediction}}\underbrace{\prod_{t=T,\ldots,1}{p(\mathbf{y}_{t-1}|\mathbf{y}_t,\mathbf{w}^x)}}_{\text{progressive full-context diffusion}} \\
&= {\int}_{\mathbf{y}_{T},\ldots,\mathbf{y}_0}{\prod_{i=1,\ldots, n}p(\mathbf{w}^y_i|\mathbf{y}_{0,i},\mathbf{w}^x)}\prod_{t=T,\ldots,1}\prod_{i=1,\ldots,n}{p(\mathbf{y}_{t-1,i}|\mathbf{y}_t,\mathbf{w}^x)}
\end{align}
where $p(\mathbf{y}_{t-1}|\mathbf{y}_t, \mathbf{w}^x)$ describes the diffusion process on contiguous representations $\mathbf{y}$.
The total number of diffusion steps is denoted as $T$.
Thereafter we omit the independent decomposition on $\mathbf{w}^y$ and $\mathbf{y}_t$. 
To apply diffusion models on discrete space, the rounding operation in \model maps the continuous vectors $\mathbf{y}$ to discrete $\mathbf{w}^y$ for each time step $t$, we hence in addition introduce both contiguous feature $\mathbf{y}$ and the discrete text $\mathbf{w}^y$ to represent the discrete text into Eq.~(\ref{eq:diffusion2}):
% One big challenge of diffusion models to learn Eq.~\ref{eq:diffusion} is that diffusion step is hard to perform on discrete space as text.
% We notice that diffusion models in Eq.~\ref{eq:diffusion} can be viewed as fully NAR models if we treat the whole diffusion process on continuous features $\mathbf{y}$ as a complete NAR models.
% To connect Eq.~\ref{eq:iter-nar} and Eq.~\ref{eq:diffusion}, \model introduces both of continuous feature $\mathbf{y}$ and the discrete text $\mathbf{w}^y$ by modeling:

\begin{align}
% \begin{split}
    p(\mathbf{w}^y|\mathbf{w}^x)
    &= {\sum_{\mathbf{w}^y_T,\ldots,\mathbf{w}^y_1}\int_{\mathbf{y}_{T},\ldots,\mathbf{y}_0}}p(\mathbf{w}^y_T|\mathbf{y}_T,\mathbf{w}^x)\prod_{t=T-1,\ldots,0}{p(\mathbf{w}^y_t|\mathbf{y}_t,\mathbf{w}^x)p(\mathbf{y}_t|\mathbf{w}^y_{t+1})} \label{eq:latent-contiguous-discrete-diffusion2} \\
    &= {\sum_{\mathbf{w}^y_T,\ldots,\mathbf{w}^y_1}\int_{\mathbf{y}_{T},\ldots,\mathbf{y}_0}}{p(\mathbf{w}^y|\mathbf{y}_{0},\mathbf{w}^x)}\prod_{t=T,\ldots,1}{p(\mathbf{y}_{t-1}|\mathbf{w}^y_t)p(\mathbf{w}^y_{t}|\mathbf{y}_t,\mathbf{w}^x)} \label{eq:latent-contiguous-discrete-nar2} \\
    &= {\int}_{\mathbf{y}_{T},\ldots,\mathbf{y}_0}{p(\mathbf{w}^y|\mathbf{y}_{0},\mathbf{w}^x)}\sum_{\mathbf{w}^y_{T},\ldots,\mathbf{w}^y_1}\prod_{t=T,\ldots,1}{p(\mathbf{y}_{t-1}|\mathbf{w}^y_t)p(\mathbf{w}^y_{t}|\mathbf{y}_t,\mathbf{w}^x)} \\
    &= {\int}_{\mathbf{y}_{T},\ldots,\mathbf{y}_0}{p(\mathbf{w}^y|\mathbf{y}_{0},\mathbf{w}^x)}\prod_{t=T,\ldots,1}\sum_{\mathbf{w}^y_t}{p(\mathbf{y}_{t-1}|\mathbf{w}^y_t)p(\mathbf{w}^y_{t}|\mathbf{y}_t,\mathbf{w}^x)}
% \end{split}
\end{align}
By rearranging Eq.~(\ref{eq:latent-contiguous-discrete-diffusion2}) and Eq.~(\ref{eq:latent-contiguous-discrete-nar2}), we can see that \model can be seen as a more generalized form of iter-NAR before marginalizing out $\{\mathbf{y}_{T},\ldots,\mathbf{y}_0\}$,
where Eq.~(\ref{eq:latent-contiguous-discrete-diffusion2}) and Eq.~(\ref{eq:latent-contiguous-discrete-nar2}) are equivalent with different computation order, 
despite the different initialization of $\mathbf{y}_T$. For NAR models, $\mathbf{y}_T$ is uniform copied from the source sentence or \emph{unk}'s token embedding~\citep{gu2017non}; for diffusion models, $\mathbf{y}_T$ is sampled from normal distribution $\mathcal{N}(0,\mathbf{I})$.
% From Eq.~\ref{eq:latent-contiguous-discrete-nar} and Eq.~\ref{eq:latent-contiguous-discrete-diffusion} of \model, we derive both of iterartive NAR model~(Eq.~\ref{eq:iter-nar}) and vanilla diffusion process~(Eq.~\ref{eq:diffusion}), by marginalizing $\mathbf{y}$ and $\mathbf{w}^y$ respectively~(detailed derivation is shown in Appendix~\ref{sec:nar-to-diffusion}).
% We argue that a key point to learn diffusion process in discrete space is leveraging an step-wise alternation process of rounding $p(\mathbf{w}^y_t|\mathbf{y}_t,\mathbf{w}^x)$ and noisy embedding $p(\mathbf{y}_{t-1}|\mathbf{w}^y_t)=\mathcal{N}(\mathbf{y}_{t-1};\sqrt{1-\beta_t}\textsc{Emb}(\mathbf{w}^y_t), \beta_t \mathbf{I})$.

% For conciseness, we intuitively~\jiangtao{it is a weak claim and is conflicted to word ``theorectical'' in the other sections. it is a safe claim and is okay to me. just make sure every part of the paper are consistent.} connect AR, interative-NAR and \model by regarding them as generation process along with different steps, i.e., token-level time steps, iteration steps, diffusion steps respectively.

\begin{figure}[t]
    \centering
    \includegraphics[width=0.7\linewidth]{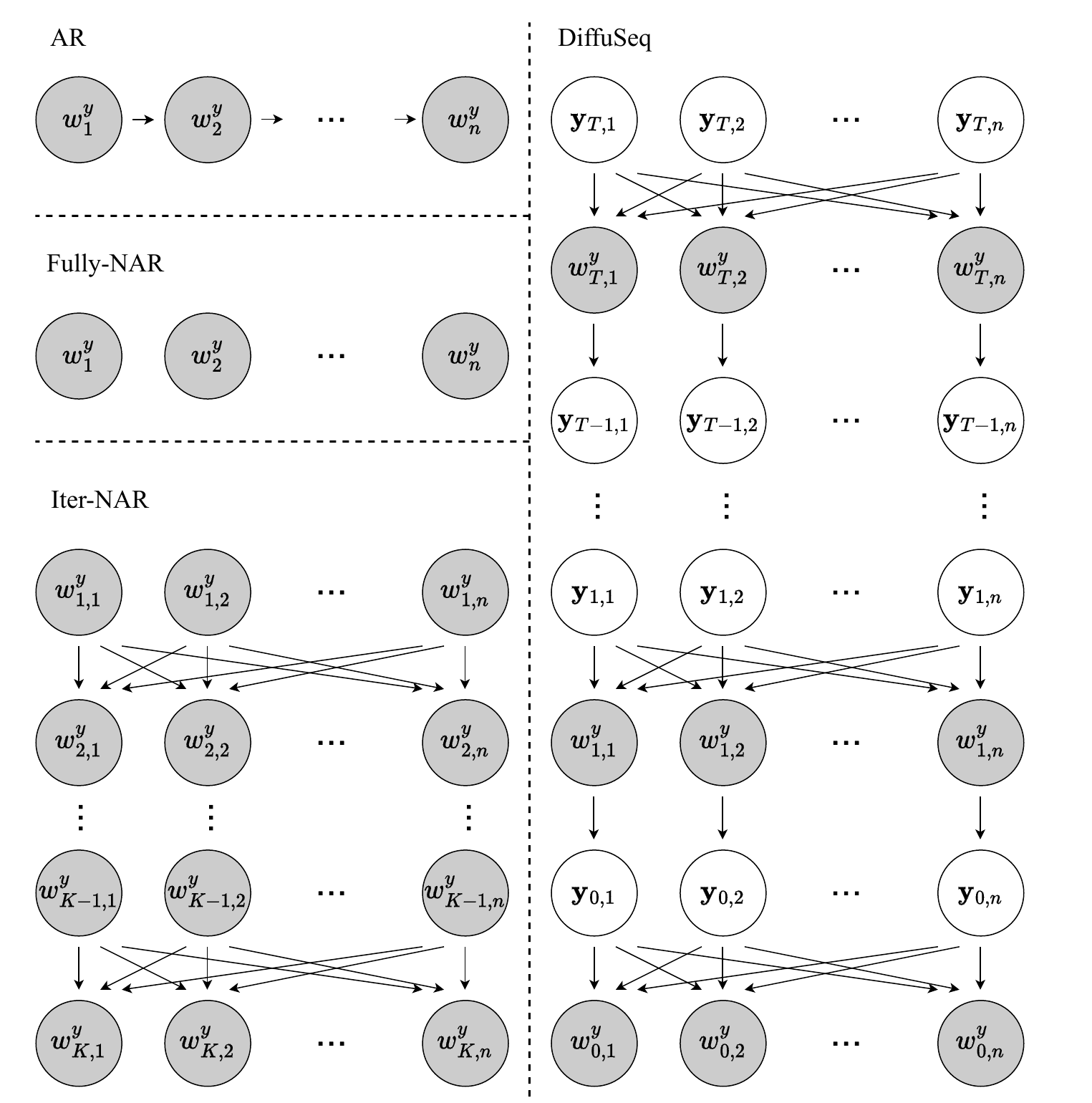}
    \caption{Graphical Models of AR, Fully NAR, iterative NAR and \textsc{DiffuSeq} models. For simplicity, we omit source node $\mathbf{w}^x$. Gray nodes indicate dependency on the source node while white nodes are independent to the source node.}
    \label{fig:graph}
\end{figure}

It is notable that unlike AR and fully NAR models generating text all at once, iterative NAR and diffusion models feature a self-corrected text generation process. The graphical comparison is shown in Figure~\ref{fig:graph}.

\label{sec:graph}

\section{From \model to Iterative NAR and Diffusion Models}
\label{sec:nar-to-diffusion}

\paragraph{From \model to Iterative NAR}
We show how to derive \model to iterative non-autoregressive model on discrete space.

\begin{align*} \\
&~p_{\model}(\mathbf{w}^y|\mathbf{w}^x) \\
=& {\sum_{\mathbf{w}^y_T,\ldots,\mathbf{w}^y_1}\int_{\mathbf{y}_{T},\ldots,\mathbf{y}_0}}{p(\mathbf{w}^y|\mathbf{y}_{0},\mathbf{w}^x)}\prod_{t=T,\ldots,1}{p(\mathbf{y}_{t-1}|\mathbf{w}^y_t)p(\mathbf{w}^y_{t}|\mathbf{y}_t,\mathbf{w}^x)} \\
=& {\sum_{\mathbf{w}^y_T,\ldots,\mathbf{w}^y_1}\int_{\mathbf{y}_{T},\ldots,\mathbf{y}_0}}p(\mathbf{w}^y_T|\mathbf{y}_T,\mathbf{w}^x)\prod_{t=T-1,\ldots,0}{p(\mathbf{w}^y_t|\mathbf{y}_t,\mathbf{w}^x)p(\mathbf{y}_t|\mathbf{w}^y_{t+1})}  && \textit{reorder computation} \\
=&\sum_{\substack{\mathbf{w}^y_T,\ldots,\mathbf{w}^y_1}}p(\mathbf{w}^y_T|\mathbf{y}_T,\mathbf{w}^x)\prod_{t=T-1,\ldots,0}{ {\int}_{\mathbf{y}_t}p(\mathbf{w}^y_t|\mathbf{y}_t,\mathbf{w}^x)p(\mathbf{y}_t|\mathbf{w}^y_{t+1})} \\
=&\sum_{\substack{\mathbf{w}^y_T,\ldots,\mathbf{w}^y_1}}p(\mathbf{w}^y_T|\mathbf{y}_T,\mathbf{w}^x)\prod_{t=T-1,\ldots,0}{p(\mathbf{w}^y_t|\mathbf{w}^y_{t+1},\mathbf{w}^x))} && \textit{marginalize over~} \mathbf{y}\\
=&\sum_{\mathbf{w}^y_1,\ldots,\mathbf{w}^y_{K-1}}{p(\mathbf{w}^y_1|\mathbf{w}^x)\vphantom{\prod_{k=1\ldots K-1}}\prod_{k=1\ldots K-1}{p(\mathbf{w}^y_{k+1}|\mathbf{w}^y_k,\mathbf{w}^x)}} && \textit{align $t$ and $k$ reversely.} \\
=&~p_{\text{iter-NAR}}(\mathbf{w}^y|\mathbf{w}^x)\\
\end{align*}

\paragraph{From \model to diffusion model}
We show how to derive \model to the straight-forward diffusion model on continuous space.

\begin{align*}
&~p_\model(\mathbf{w}^y|\mathbf{w}^x) \\
=& {\sum_{\mathbf{w}^y_T,\ldots,\mathbf{w}^y_1}\int_{\mathbf{y}_{T},\ldots,\mathbf{y}_0}}{p(\mathbf{w}^y|\mathbf{y}_{0},\mathbf{w}^x)}\prod_{t=T,\ldots,1}{p(\mathbf{y}_{t-1}|\mathbf{w}^y_t)p(\mathbf{w}^y_{t}|\mathbf{y}_t,\mathbf{w}^x)} \\
=& {\int}_{\substack{\mathbf{y}_{T},\ldots,\mathbf{y}_0}}{p(\mathbf{w}^y|\mathbf{y}_{0},\mathbf{w}^x)}\prod_{t=T,\ldots,1}{\sum_{\mathbf{w}^y_t}p(\mathbf{y}_{t-1}|\mathbf{w}^y_t)p(\mathbf{w}^y_{t}|\mathbf{y}_t,\mathbf{w}^x)} \\
=& {\int}_{\mathbf{y}_{T},\ldots,\mathbf{y}_0}{p(\mathbf{w}^y|\mathbf{y}_{0},\mathbf{w}^x)}\prod_{t=T,\ldots,1}{p(\mathbf{y}_{t-1}|\mathbf{y}_t,\mathbf{w}^x)} && \textit{marginalize over~}\mathbf{w}^y \\
=&~p_{\text{diffusion}}(\mathbf{w}^y|\mathbf{w}^x)
\end{align*}

\section{Details of Experiments}
\subsection{Processing of Question Generation Dataset}
\label{sec:appendix-qg}
To construct high-quality document-question pairs from the Quasar-T dataset, which consists of $\langle{document}_i, {question}, {answer}\rangle$ triplets, we extract $\langle{document}_i, question\rangle$ pairs if ${answer}$ exactly matches ${document}_i$. After pre-processing, we obtain 119K document-question training pairs.

\subsection{Settings of Baselines}
\label{sec:appendix-basline}
We compare the settings of different models, including the number of parameters and how to sample the different output sentences, as shown in~\tabref{tb:comp-models}. For plain GRU-based encoder-decoder methods, we do not implement diversity search algorithms on it, thus its sentence-level diversity could be very poor. For NAR-LevT, we set max iteration to 9 and follow the termination condition mentioned in the original paper. For GPVAE-T5, we tune the scalar to find the best trade-off between quality and diversity on the dev set. The scalars of all four tasks are set to 2. We implement GPT2 baselines using HuggingFace \texttt{Transformers} and for the baseline Transformer-base, we use \texttt{Fairseq}.

\begin{table}[!th]
\centering
\begin{threeparttable}[b]
\caption{The comparison for different models}
\label{tb:comp-models}
\begin{tabular}{l|lll}
\toprule
Models & \# Parameters & Learning Paradigm & Diversity Source \\
\midrule
GRU-attention & 65M & encoder-decoder & - \\
Transformer-base & 80M & encoder-decoder & Temperature/DBS \\
\midrule
GPT2-base FT & 117M & pretrain-finetune & Hybrid strategy\tablefootnote{Including top-p sampling, temperature, diversity beam search (DBS) and etc. Implement using HuggingFace \texttt{Transformers} \url{https://github.com/huggingface/transformers}} \\
GPT2-large FT & 774M & pretrain-finetune & Hybrid strategy \\
GPVAE-T5 & 220M & pretrain+VAE & Gaussian sampling \\
\midrule
NAR-LevT & 80M & non-autoregressive & - \\
\model & 91M & non-autoregressive & Gaussian sampling \\
\bottomrule
\end{tabular}
\end{threeparttable}
\end{table}

\subsection{Diversity and quality Trade-off Settings}
\label{sec:appendix-trade-off}
We list the details to obtain~\Figref{fig:trade_off_qual_div}.
For GPVAE-T5, we set different scalars as $1,2,3,4$. For \model, we choose trained models at different training steps to achieve different trade-off points. For other baselines, there is no explicit factor to control the diversity generation, so we leave them as single points in the figure.

\subsection{Metrics}
\label{sec:appendix-metrcis}
The used BLEU score is sentence-level smoothed from BLEU-1 to 4, and used ROUGE-L score is longest common subsequence based statistics. The implementation is based on \texttt{NLTK}\footnote{\url{https://www.nltk.org/_modules/nltk/translate/bleu_score.html}} and \texttt{torchmetrics}.
The n-gram based metrics may fail to capture the semantic meaning of sentences, so we consider using BERTScore\footnote{\url{https://github.com/Tiiiger/bert_score}}. Specifically, we use \texttt{microsoft/deberta-xlarge-mnli} to help BERTScore correlate better with human scores.

\subsection{Generation Results}
For different tasks we list some generation examples. As we can see in \tabref{tb:case cc}, \tabref{tb:case qg} and \tabref{tb:case ts}, \model tends to generate diverse outputs, but sometimes the sentence is not as fluent as finetuned GPT2.
\label{sec:appendix-case}
\begin{table}[!th]
\centering
\begin{threeparttable}[b]
\caption{Sample outputs with different random seed in Dialogue test set.}
\label{tb:case cc}

\begin{tabular}{l|l}
\toprule
\multicolumn{2}{l}{\textit{\textbf{Utterance}: How long does the dye last?}}\\
\multicolumn{2}{l}{\textit{\textbf{Response}: Just did this two days ago, not sure how it'll fade yet!}}\\
\midrule
\textbf{GPVAE-T5} &
\textbf{NAR-LevT} \\
 & \\
 * \makecell[l]{I'm not sure, I'm not sure. I've tested it a few \\times, but I don't know for sure. I've}&* half .  \\
 & \\

 * \makecell[l]{I'm not sure. I'm not sure how long it lasts, I'm \\sure it 'll get better. It's been a while since} & * half .\\
 & \\
 * \makecell[l]{I've been using it for about a year and a half.\\ I've been using it for about a year and a half.} & * half . \\
\midrule
% \midrule
\textbf{GPT2-large finetune} & \textbf{\model} \\
 & \\
* Two weeks in my case. & * About an hour, 5 days or so. \\ 
 & \\
* I've had it for about a year.  & * 4 days.\\ 
 & \\
* \makecell[l]{The dye can sit around for a month then you \\can wash it.} & * \makecell[l]{I'm not sure about this, about the same \\kind of time.} \\

\bottomrule
\end{tabular}
\end{threeparttable}
\end{table}

\begin{table}[!th]
\centering
\begin{threeparttable}[b]
\caption{Sample outputs with different random seed in Question Generation test set.}
\label{tb:case qg}

\begin{tabular}{l|l}
\toprule
\multicolumn{2}{l}{\makecell[l]{\textit{\textbf{Statement}: The Japanese yen is the official and only currency recognized in Japan.}}}\\
\multicolumn{2}{l}{\textit{\textbf{Question}: What is the Japanese currency?}}\\
\midrule
\textbf{GPVAE-T5} & \textbf{NAR-LevT} \\
* What is the japanese currency &  * What is the basic unit of currency for Japan ? \\
* What is the japanese currency &  * What is the basic unit of currency for Japan ? \\
* What is the japanese currency &  * What is the basic unit of currency for Japan ? \\

\midrule
% \midrule
\textbf{GPT2-large finetune} & \textbf{\model} \\
* What is the basic unit of currency for Japan? & * What is the Japanese currency \\ 
* What is the Japanese currency  & * Which country uses the ``yen yen'' in currency\\ 
* What is the basic unit of currency for Japan? & * What is the basic unit of currency? \\
\bottomrule
\end{tabular}
\end{threeparttable}
\end{table}

% \subsection{Ablation: Jointly Training}
% \label{sec:appendix-ablation}

\begin{table}[!th]
\centering
\begin{threeparttable}[b]
\caption{Sample outputs with different random seed in Text Simplification test set.}
\label{tb:case ts}

\begin{tabular}{l|l}
\toprule
\multicolumn{2}{l}{\makecell[l]{\textit{\textbf{Complex sentence}: People can experience loneliness for many reasons, and many life events} \\ \textit{\quad \quad may cause it, such as a lack of friendship relations during childhood and adolescence, or } \\ \textit{\quad \quad the physical absence of meaningful people around a person.}}}\\
\multicolumn{2}{l}{\textit{\textbf{Simplified}: One cause of loneliness is a lack of friends during childhood and teenage years.}}\\
\midrule
\textbf{GPVAE-T5} & \textbf{NAR-LevT} \\
 & \\
* \makecell[l]{People can experience loneliness for many \\reasons, and many life events may cause it, \\such as a lack of friendship relations during \\childhood and adolescence, or the physical \\absence of meaningful people around a person} & * \makecell[l]{People may experience  reashapphap-\\phapphapphapphappabout life reasit.} \\
 & \\
* \makecell[l]{People can experience loneliness for many \\reasons, and many life events may cause it, \\such as a lack of friendship relations during \\childhood and adolescence, or the physical \\absence of meaningful people around a person} & * \makecell[l]{People may experience  reashapphap-\\phapphapphapphappabout life reasit.} \\
 & \\
* \makecell[l]{People can experience loneliness for many \\reasons, and many life events may cause it, \\such as a lack of friendship relations during \\childhood and adolescence, or the physical \\absence of meaningful people around a person} & * \makecell[l]{People may experience  reashapphap-\\phapphapphapphappabout life reasit.} \\
\midrule
% \midrule
\textbf{GPT2-large finetune} & \textbf{\model} \\
 & \\
 * Loneliness can be caused by many things. & * Many life events may cause of loneliness \\ 
  & \\
* Loneliness can affect people in many ways.  & * \makecell[l]{People can also be very experience \\loneliness for many reasons.}\\ 
 & \\
* Loneliness can be caused by many things. & * \makecell[l]{People can experience loneliness for many \\reasons, and many life events may, cause it.} \\
\bottomrule
\end{tabular}
\end{threeparttable}
\end{table}

\end{document}